\documentclass[twoside]{article}
\usepackage[accepted]{claiunconf2023}

\usepackage[round]{natbib}

\usepackage{graphicx} % For inserting images
\usepackage{amsmath}
\usepackage{amssymb}
\usepackage{enumitem} % For controlling separation in lists
\usepackage{booktabs} % For professional-quality tables
\usepackage[T1]{fontenc}

\usepackage[figuresright]{rotating}
\usepackage{multirow}
\usepackage{dblfloatfix}   
\usepackage[bottom]{footmisc}

\usepackage{hyperref}

\usepackage{algorithm}% http://ctan.org/pkg/algorithm
\usepackage{algpseudocode}% http://ctan.org/pkg/algorithmicx
\newcommand{\algrule}[1][.4pt]{\par\vskip.3\baselineskip\hrule height #1\par\vskip.3\baselineskip}

\usepackage{array}
\newcolumntype{P}[1]{>{\centering\arraybackslash}p{#1}}

\usepackage[dvipsnames,table]{xcolor}
\usepackage{pifont}

\renewcommand\fbox{\fcolorbox{orange}{white}}

\let\oldpm\pm
\renewcommand{\pm}{\scriptstyle\,\oldpm\,}

\newcommand{\eragem}{ER\,+\,\mbox{A-GEM} }
\newcommand{\ergem}{ER\,+\,\mbox{GEM} }
\newcommand{\jointgem}{Joint\,+\,\mbox{GEM} }
\newcommand{\jointagem}{Joint\,+\,\mbox{A-GEM} }

\newcommand{\cc}{\cellcolor{gray!20}}

\begin{document}

\runningtitle{Two Complementary Perspectives to Continual Learning}

%%%%%%%%%%%%%%%%%%%%%%%%%%%%%%%%%%%%%%%%%%%%

\twocolumn[

\CLAIUnconftitle{Two Complementary Perspectives to Continual Learning: \\Ask Not Only \emph{What} to Optimize, But Also \emph{How}}

\CLAIUnconfauthor{Timm Hess \And Tinne Tuytelaars \And Gido M. van de Ven}

\CLAIUnconfaddress{KU Leuven \And KU Leuven \And KU Leuven} ]

\begin{abstract}
    Recent years have seen considerable progress in the continual training of deep neural networks, predominantly thanks to approaches that add replay or regularization terms to the loss function to approximate the joint loss over all tasks so far. However, we show that even with a perfect approximation to the joint loss, these approaches still suffer from temporary but substantial forgetting when starting to train on a new task. Motivated by this `stability gap', we propose that continual learning strategies should focus not only on the optimization objective, but also on the way this objective is optimized. While there is some continual learning work that alters the optimization trajectory (e.g., using gradient projection techniques), this line of research is positioned as alternative to improving the optimization objective, while we argue it should be complementary. 
    In search of empirical support for our proposition, we perform a series of pre-registered experiments combining replay-approximated joint objectives with gradient projection-based optimization routines. %. Although these experiments fail to show clear and consistent benefits, our conceptual arguments as well as some of our empirical results still highlight the distinctive importance of the optimization trajectory in continual learning, thereby opening up a new direction for continual learning research.
    However, this first experimental attempt fails to show clear and consistent benefits. Nevertheless, our conceptual arguments, as well as some of our empirical results, demonstrate the distinctive importance of the optimization trajectory in continual learning, thereby opening up a new direction for continual learning research.
    %We hope that our conceptual arguments, together with some of our empirical results, inspire and inform further research aimed at developing novel optimization routines for continual learning.
\end{abstract}

%%%%%%%%%%%%%%%%%%%%%%%%%%%%%%%%%%%%%%%%%%%%

\section{INTRODUCTION}
Learning continually from a stream of non-stationary data is challenging for deep neural networks. When these networks are trained on something new, their default behaviour is to quickly forget most of what was learned before \citep{mccloskey1989catastrophic,ratcliff1990connectionist}. Considerable progress has been made in recent years towards overcoming such `catastrophic forgetting', for a large part thanks to methods using replay \citep{robins1995catastrophic,rolnick2019experience} and regularization \citep{kirkpatrick2017overcoming,li2017learning}. These methods work by adding extra terms to the loss function, and they can be interpreted as attempts to approximate the joint loss over all tasks so far.

Recently a peculiar property of replay and regularization methods was pointed out. These approaches tend to suffer from substantial forgetting when starting to learn a new task, although this forgetting is often temporary and followed by a phase of performance recovery \citep{delange2023continual}. We postulate that avoiding this `stability gap' is important, both because the transient drops in performance themselves can be problematic (e.g., for safety-critical applications) and because doing so might lead to more efficient and better performing algorithms, as constantly having to re-learn past tasks seems wasteful. 
Importantly, however, we demonstrate that the stability gap cannot be overcome by merely improving replay or regularization. This motivates us to propose that, instead, continual learning needs an additional perspective: rather than focusing only on \emph{what} to optimize (i.e., the optimization objective), the field should also think about \emph{how} to optimize (i.e., the optimization trajectory).
%In particular, we believe optimal for continual learning wout be to optimize a suitable proxy for the joint loss in a way that does not (strongly) interfere with the loss on past tasks.

There are existing works that explore modifying the optimization trajectory as a mechanism for continual learning, but so far this line of research has been positioned as alternative to improving the optimization objective, rather than as complementary. A prime example is Gradient Episodic Memory (GEM; \citealp{lopez2017gradient}), which alters the optimization process by projecting gradients to encourage parameter updates that do not strongly interfere with old tasks. Crucially, GEM applies this optimization routine to optimizing the loss on the new task, while according to our proposal such modified optimization routines should be used to optimize approximations of the joint loss. As a first evaluation of the merits of our proposition, in this work we use GEM's gradient projection-based optimization routine to instead optimize replay-approximated versions of the joint loss. In a series of pre-registered experiments, using both domain- and class-incremental learning benchmarks, we test whether this combined approach reduces the stability gap, and whether this in turn leads to higher learning efficiency and better final performance.

The remainder of the paper is organized as follows. In section~\ref{sec:new_perspective} we develop our main proposition. In section~\ref{sec:constrained_optimization} we review existing optimization-based methods for continual learning. In section~\ref{sec:method} we propose experiments of which we hope that they can provide proof-of-concept demonstrations for our main proposition, and in section~\ref{sec:experimental_details} we describe the detailed pre-registered experimental protocol. In section~\ref{sec:results} we present the results, and we end with a discussion and outlook in section~\ref{sec:discussion}.

%%%%%%%%%%%%%%%%%%%%%%%%%%%%%%%%%%%%%%%%%%%%

\section{TWO PERSPECTIVES TO CONTINUAL LEARNING}
\label{sec:new_perspective}
In this section we use conceptual arguments and preliminary data to develop the  proposition that continual learning should focus not only on \emph{what} to optimize, but also~on~\emph{how}. We first describe the current dominant approach to continual learning (subsection~\ref{sec:dominant}), we then point out a fundamental issue with this approach (subsection~\ref{sec:stability_gap}), and we finally propose a complementary approach and explain why it could address this issue (subsection~\ref{sec:alternative}).

To help us reason about the different approaches that continual learning methods could take, in this section we consider the following continual learning problem.
Assume a model~$f_w$, parameterized by $w$, that has learned a set of weights~$\widehat{w}_{\text{old}}$ for an initial task\footnote{The term `task' is used here in a rather general way; it loosely refers to a combination of a data distribution and a loss function.}, or a set of tasks, by optimizing a loss function~$\ell_{\text{old}}$ on training data~$D_{\text{old}}\sim\mathcal{D}_{\text{old}}$. We then wish to continue training the same model on a new task, by optimizing a loss function~$\ell_{\text{new}}$ on training data~$D_{\text{new}}\sim\mathcal{D}_{\text{new}}$, in such a way that the model maintains (or possibly improves) its performance on the previously learned task(s). As has been thoroughly described in the continual learning literature, if the model is trained on the new task in the standard way (i.e., optimize the new loss~$\ell_{\text{new}}$ with stochastic gradient descent), the typical result is catastrophic forgetting and a solution~$\widehat{w}_{\text{new}}$ that is good for the new task but no longer for the old one(s).

\subsection{The Standard Approach to Continual Learning: Improving the Loss Function}
\label{sec:dominant}
To mitigate catastrophic forgetting, continual learning research from the past few years has typically focused on making changes to the loss function that is optimized. In particular, rather than optimizing the loss on the new task, many continual learning methods can be interpreted as optimizing an approximate version of the joint loss:
\begin{equation}
    \label{eq:approx_joint}
    \widetilde{\ell}_{\text{joint}} = \ell_{\text{new}} + \widetilde{\ell}_{\text{old}},
\end{equation}
with $\widetilde{\ell}_{\text{old}}$ the method's proxy for the loss on the old tasks.

A straight-forward example of this approach is `experience replay', which approximates $\ell_{\text{old}}$ by revisiting a subset of previously observed examples that are stored in an auxiliary memory buffer. In continual learning experiments, typically limits are imposed on the buffer's storage capacity and/or on the computational budget for training the model \citep{lesort2020continual,wang2022memory,prabhu2023computationally}. Both constraints prevent full replay of all previously observed data, meaning that $\ell_{\text{old}}$ can only be approximated. A wide range of studies aims to improve the quality of this approximation, for example by modifying the way samples are selected to be stored in the buffer \citep{rebuffi2017icarl,chaudhry2019tiny,aljundi2019gradient,lin2021clear,mundt2023wholistic}, or by adaptively selecting which samples from the buffer to replay \citep{riemer2018learning,aljundi2019mir}. As an alternative to storing past samples explicitly, generative models can be learned to approximate the input distributions of previous tasks \citep{robins1995catastrophic, shin2017continual, van2020brain}.

Another popular class of methods for continual learning is based on regularization. As proxy for the loss on the old tasks, these methods add regularizing terms to the loss that impose penalties either for changes to the network's weights (`parameter regularization'; \citealp{kirkpatrick2017overcoming,zenke2017continual,aljundi2018memory}) or for changes to the network's input-output mapping (`functional regularization'; \citealp{li2017learning,dhar2019learning,lee2019overcoming,Titsias2020Functional}). That these methods can be interpreted as attempts to approximate the joint loss can be shown by taking either a Bayesian perspective \citep{nguyen2018variational,farquhar2019unifying,kao2021natural,rudner2022continual} or a geometric perspective \citep{Kolouri2020Sliced}.

In summary, both replay- and regularization-based methods for continual learning operate by changing the loss function that is optimized, often with the aim of creating an approximate version of the joint loss. When developing new continual learning methods of this kind, the challenge is to design better objective functions.

\subsection{The Stability Gap: A Challenge for the Standard Approach to Continual Learning}
\label{sec:stability_gap}
It was recently pointed out that even when replay- or regularization-based methods are considered to perform well (in the sense that they obtain good performance on both old and new tasks after finishing training on the new task), these methods still suffer from substantial, albeit often temporary, forgetting during the initial phase of training on a new task \citep{delange2023continual}. Until recently, this phenomenon -- referred to as the \emph{stability gap} -- had not been observed, or been paid little attention to, due to the common evaluation setup in continual learning that only measures performance after training on the new data has converged. Nevertheless, the stability gap is undesirable, especially for safety-critical applications in which sudden drops in performance can be highly problematic. But also from the perspective of computational efficiency, or even if only the final learning outcome is of interest, avoiding the stability gap might be beneficial, as preventing forgetting seems easier and more efficient than having to re-learn later on (see Figure 4 of Van de Ven et al.~(\citeyear{van2020brain}) for empirical support for this intuition).

\begin{figure}[!t]
    \centering
    \vskip 0.05in
    \includegraphics[width=0.46\textwidth]{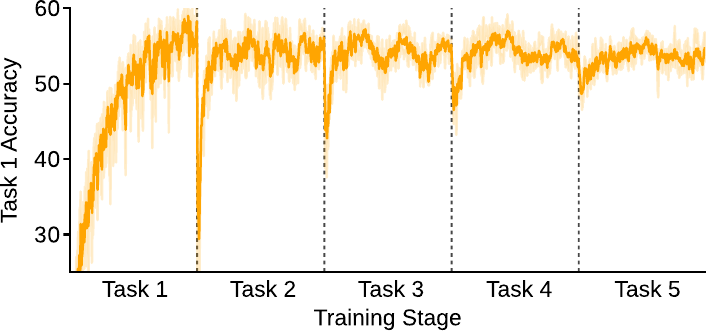}
    \vskip -0.04in
    \caption{\textbf{The stability gap occurs even with incremental joint training (or `full replay').} Shown is the test accuracy on the first task while the network is incrementally trained on all five tasks of Domain \mbox{CIFAR}. During the $n$-th task, the network is trained jointly on all training data from the first $n$ tasks. Even with this ideal approximation to $\ell_{\text{joint}}$, performance severely drops upon encountering a new task. Displayed are the means over five repetitions, shaded areas are $\pm1$ standard error of the mean. Vertical dashed lines indicate task switches.}
    \label{fig:stabilit-gap-exemplar-domcif100}
\end{figure}

Why does the stability gap happen? One possibility is that the stability gap is due to imprecision in the approximations of the joint loss made by replay and regularization. If this were the case, it could be interpreted as good news for the standard approach to continual learning, as it would imply that by continuing to improve the quality of replay or regularization the stability gap could be overcome. However, this is not the case, as in preliminary experiments we find that the stability gap is consistently observed even with incremental joint training (Figure~\ref{fig:stabilit-gap-exemplar-domcif100}). This indicates that with better approximations to the joint loss alone, the stability gap cannot be solved.

\subsection{Proposed Complementary Approach: ~~~~~ Improving the Optimization Trajectory}
\label{sec:alternative}
The above observations relating to the stability gap suggest that the standard approach to continual learning of focusing on the loss function is not sufficient.
We believe that continual learning would benefit from an additional perspective: rather than concentrating only on improving the optimization objective (i.e., what loss function to optimize), we argue that continual learning should also focus on improving the optimization trajectory (i.e., how to optimize that loss function).

\begin{figure}[!b]
    \centering
    \vskip -0.031in
    \includegraphics[width=0.41\textwidth]{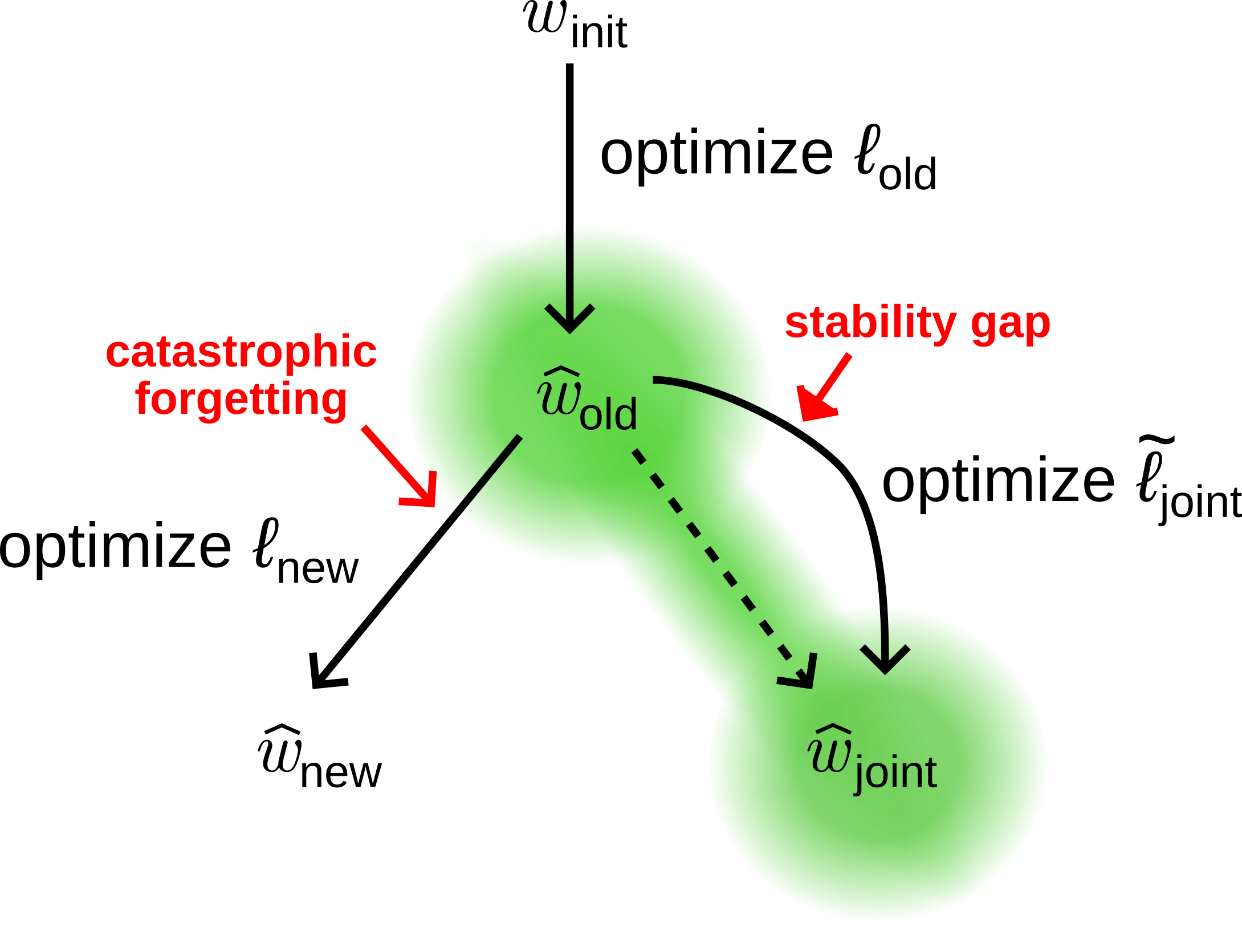}
    \vskip -0.1in
    \caption{\textbf{Schematic of the stability gap, and how adjusting the optimization trajectory could avoid it.} When, starting from a solution for the old tasks~($\widehat{w}_{\text{old}}$), a proxy of the joint loss~($\widetilde{\ell}_{\text{joint}}$) is optimized with standard stochastic gradient descent, the optimization trajectory first passes through a region in parameter space with high loss on the old tasks before converging to a solution that is good for all tasks~($\widehat{w}_{\text{joint}}$). Work on mode connectivity suggests that a low-loss path between $\widehat{w}_{\text{old}}$ and $\widehat{w}_{\text{joint}}$ exists as well (dashed arrow), indicating that it should be possible to overcome the stability gap with a different optimization routine. Green shading indicates areas of low loss on the old tasks.}
    \label{fig:stability-gap-schematic}
    \vskip -0.031in
\end{figure}

To help explain why we believe that focusing on the optimization trajectory can yield benefits, we revisit the continual learning problem discussed at the start of section~\ref{sec:new_perspective}, which is schematically illustrated in Figure~\ref{fig:stability-gap-schematic}. Starting point is a model~$f_w$ that has already learned a solution~$\widehat{w}_{\text{old}}$ by optimizing loss~$\ell_{\text{old}}$. Continuing to train this model by optimizing loss~$\ell_{\text{new}}$ would result in catastrophic forgetting. Instead, as discussed, the standard approach in continual learning is to optimize $\widetilde{\ell}_{\text{joint}} = \ell_{\text{new}} + \widetilde{\ell}_{\text{old}}$, an approximation to the joint loss, rather than $\ell_{\text{new}}$. Optimizing a suitably approximated version of the joint loss results in a solution~$\widehat{w}_{\text{joint}}$ that is good for both the old and the new tasks. However, if this loss is optimized with standard stochastic gradient descent, the trajectory that is taken from $\widehat{w}_{\text{old}}$ to $\widehat{w}_{\text{joint}}$ goes through a region in parameter space where the loss on the old tasks is high. The corresponding transient drop in performance on the old tasks is the stability gap.

A first possibility that must be dealt with is that the stability gap is unavoidable, in the sense that there simply is no path from $\widehat{w}_{\text{old}}$ to $\widehat{w}_{\text{joint}}$ that does not traverse a region where the performance on old tasks is poor. Although this is theoretically possible, this option seems unlikely given recent work on mode connectivity in deep neural networks \citep{draxler2018essentially,garipov2018loss} showing that different local optima found by stochastic gradient descent are often connected by simple paths of non-increasing loss. In particular, \cite{mirzadeh2020linear} showed that when optimizing a neural network using stochastic gradient descent on the joint loss while starting from a single task solution, the resulting joint solution is connected to the single task solution by a linear manifold of low loss on the single task. The same holds when, starting from a single task solution, the replay-approximated joint loss is optimized rather than the joint loss itself \citep{verwimp2021rehearsal}.

The above work on mode connectivity thus suggests that by changing the optimization routine it should be possible to avoid the stability gap. Besides that reducing the stability gap is important for safety-critical applications, we believe that it could bring other benefits for continual learning as well. For example, getting rid of the repeated re-learning cycles that characterize the stability gap could increase the learning efficiency. Moreover, if the loss function is non-convex, as is the case with deep neural networks, changing the way the loss is optimized could also lead to different, and hopefully better, final learning outcomes. The possibility of improved final performance is supported by the observation from \cite{caccia2022new} that the abrupt forgetting after task switches is not always recovered later on. This leads us to the following three hypotheses:

\vspace{0.065in}
\fbox{\begin{minipage}{0.96\columnwidth}
\vspace{0.04in}
~\textbf{Main hypothesis}

~Better optimization routines for continual learning can:
\hspace{1in}
\begin{itemize}[leftmargin=1.05cm,nosep]
    \item[\textbf{(H1)}] reduce the stability gap.
\end{itemize}
\vspace{0.06in}
~\textbf{Secondary hypotheses}

~Reducing the stability gap can:
\hspace{1in}
\begin{itemize}[leftmargin=1.05cm,nosep]
    \item[\textbf{(H2)}] increase learning efficiency;
    \item[\textbf{(H3)}] improve the final learning outcome.
\end{itemize}
\vspace{0.03in}
\end{minipage}}
\vspace{0.03in}

% \vspace{0.05in}
% \fbox{\begin{minipage}{\columnwidth}
% In short, we hypothesize that better optimization routines for continual learning could yield three benefits:
% \hspace{1in}
% \begin{itemize}[leftmargin=1.05cm,nosep]
%     \item[\textbf{(H1)}] Reduce the stability gap;
%     \item[\textbf{(H2)}] Increase learning efficiency;
%     \item[\textbf{(H3)}] Improve the final learning outcome.
% \end{itemize}
% \end{minipage}}
% \vspace{0.03in}

\paragraph{How to Improve Optimization for Continual Learning?}
In the above we have made an argument that continual learning should focus on improving its optimization routines, but we have not yet discussed \emph{how} this could be done. To avoid the stability gap, an optimization routine is needed that is less greedy than standard stochastic gradient descent and that favors parameter updates that do not substantially increase the loss on old tasks. Interestingly, as we review in section~\ref{sec:constrained_optimization}, optimization routines based on gradient projection have been explored in the continual learning literature that already have these properties. Importantly, however, currently these gradient projection-based optimization routines are not used in the way envisioned by us, as they are used to optimize the loss on the new task rather than an approximated version of the joint loss.
%How to design suitable optimization schemes for continual learning is largely an open research question, although some works have started to explore this direction. In section~\ref{sec:constrained_optimization} we review proposed approaches, and we discuss why so far they have not yet delivered the benefits hypothesized above.

\subsection{Another Way to Avoid the Stability Gap?}
%\subsection{Another Way to Avoid the Stability Gap: ~~~~~ Using Task-specific Components}
An alternative approach that can circumvent the stability gap in continual learning is to use certain parts of the network only for specific tasks. This approach is employed by network expansion methods \citep{rusu2016progressive,yoon2018lifelong,yan2021dynamically} and parameter isolation methods \citep{serra2018overcoming,masse2018alleviating}. To see why this approach can help to avoid the stability gap, consider the extreme case of using a separate sub-network for each task. In this case there is no forgetting at all, and thus also no stability gap. However, the use of task-specific components has some important disadvantages. Firstly, if task identity is not always provided, as is the case with domain- and class-incremental learning \citep{van2022three}, it might not be clear which parts of the network should be used. This issue could be addressed by inferring task identity, for example using generative models or other out-of-distribution detection techniques \citep{van2021class,henning2021posterior,kim2022theoretical,zajkac2023prediction}, but such task inference can be challenging. Secondly, having separate parts of the network per task limits the potential of positive transfer between tasks, which is an important desideratum for continual learning \citep{hadsell2020embracing}. %This justifies our focus on addressing the stability gap by improving the optimization routine.
%TEXT POSTED IN OUR REBUTTAL:
%Using parts of the network only for specific tasks (e.g., as done by network expansion, parameter isolation and orthogonal projection methods) is an alternative way to avoid the stability gap, but that there are two important disadvantages with this approach. First, if task-ID is not provided at test time, it might not be clear which part of the network should be used for inference; although this issue could be addressed by inferring task identity, for example using OOD-based approaches. Second, having separate parts of the network per task limits the potential of positive transfer between tasks, which is an important desiderata for continual learning.

%%%%%%%%%%%%%%%%%%%%%%%%%%%%%%%%%%%%%%%%%%%%

\section{GRADIENT PROJECTION-BASED OPTIMIZATION}
\label{sec:constrained_optimization}

A tool that has been explored in the continual learning literature for modifying the way a given loss function~$\ell(w)$ is optimized is `gradient projection'. With gradient projection, rather than basing the parameter updates on the original gradient $g=\nabla_w \ell(w)$, they are based on a projected version $\bar{g}$ of that gradient. Important from the perspective of our paper, gradient projection does not alter the loss function that is optimized (e.g., the loss landscape and its local minima remain unchanged), it only changes the way the loss function is optimized \citep{kao2021natural}. In this section we review two current lines of continual learning studies that make use of gradient projection.

\subsection{Orthogonal Gradient Projection}
Orthogonal gradient projection methods aim to avoid interference between tasks by confining the training of each new task to previously unused subspaces. To restrict parameter updates to directions that do not interfere with the performance on old tasks, the gradient of the loss on the new task is projected to the orthogonal complement of the `gradient subspaces' of old tasks. Various ways to construct these gradient subspaces have been proposed.
Several studies use the subspaces spanned by all layer-wise inputs of old tasks, which they characterize using conceptors \citep{he2018overcoming} or by iteratively accumulating projector matrices using a recursive least squares algorithm \citep{zeng2019continual,guo2022adaptive}.
%\cite{zeng2019continual} and \cite{guo2022adaptive} use the subspaces spanned by all layer-wise inputs of old tasks, which they track by iteratively accumulating a projector matrix $P$ using a recursive least squares algorithm.
To reduce the memory and computational costs, \cite{saha2021gradient} approximate the input subspaces of each task using singular value decomposition and its $k$-rank approximation. \cite{farajtabar2020orthogonal} instead use the span of a set of stored gradient directions of old tasks as the gradient subspace to protect.

Constraining sequential optimization to orthogonal subspaces can mitigate forgetting effectively, but it restricts the learning of new tasks to successively smaller subspaces and it eliminates the potential to further improve the model with respect to old tasks. 
Orthogonal gradient projection methods especially struggle or break down when the input spaces of different tasks substantially overlap with each other \citep{he2018continual}.

Recent advancements in this line of work therefore focus on relaxing the constraints of the orthogonal projection framework to enable knowledge transfer between tasks. \cite{deng2021flattening} dynamically scale gradients and search for flatter minima, and \cite{lin2022trgp} use the idea of `trust regions' \citep{schulman2015trust} to selectively relax constraints for protected subspaces of old tasks most related to the new task. As an alternative, \cite{kao2021natural} take a Bayesian perspective and transform the gradients using the inverse of a Kronecker-factored approximation to the Fisher information matrix, and additionally protect previous knowledge by parameter-based regularization.

\subsection{Gradient Episodic Memories}
\label{sec:gem}
Another class of gradient projection-based optimization methods for continual learning, which also does not enforce strict orthogonality of future updates, is based on Gradient Episodic Memory~(GEM; \citealp{lopez2017gradient}). The projection mechanism of this approach is motivated by a constrained optimization problem, where the goal is to optimize $\ell_{\text{new}}$ without increasing $\ell_{\text{old}}$: 
\begin{equation}
    \text{min}_w \ \ell_{\text{new}}(w), ~~\text{such that} \ \ell_{\text{old}}(w) \leq \ell_{\text{old}}(\widehat{w}_{\text{old}}).
\end{equation}
To determine whether a parameter update based on $g=\nabla_w \ell_{\text{new}}(w)$ might increase $\ell_{\text{old}}$, the gradient(s) for the old task(s) are estimated using examples from a replay buffer: $g_{\text{old}} = \nabla_w \widetilde{\ell}_{\text{old}}(w)$. If the directions of $g$ and $g_{\text{old}}$ align (in the sense that their angle does not exceed $90^\circ$),
%(in the sense that their angle is $\leq 90^\circ$), 
it is conjectured that a parameter update based on $g$ is unlikely to increase $\ell_{\text{old}}$, and $g$ is left unchanged. %If $\langle g, \widetilde{g}_{old}\rangle < 0$, $g$ is projected to
If the angle between $g$ and $g_{\text{old}}$ exceeds $90^\circ$, $g$ is projected to 
$\bar{g}=g-\frac{g^{\text{T}}g_{\text{old}}}{g_{\text{old}}^{\text{T}}g_{\text{old}}}g_{\text{old}}$,
which is the closest gradient to $g$ (in $l_2$-norm) with a $90^\circ$~angle to $g_{\text{old}}$.
%, such that $\langle \bar{g}, \widetilde{g}_{old}\rangle \geq 0$. 
%that is withing the angle-bound with respect to $g_{\text{old}}$.
%has a non-negative angle with $g_{\text{old}}$.
Because in our continual learning example it is the case that $\ell_{\text{old}}$ encompasses all past tasks, this description actually corresponds to Averaged GEM~(\mbox{A-GEM}; \citealp{chaudhry2018efficient}), a computationally more efficient version of GEM. The original formulation of GEM enforces $\bar{g}$ to align with the gradient of each individual past task \citep{lopez2017gradient}, see Appendix~\ref{sec:gem_memory_strength} for details.

Given that GEM and A-GEM explicitly aim to prevent increases of the loss on old tasks, these methods might be able to avoid the stability gap. Empirically, however, this is not the case, as GEM suffers from considerably larger stability gaps than experience replay \citep{delange2023continual}. Moreover, also in terms of final performance, experience replay consistently outperforms both GEM and \mbox{A-GEM} \citep{de2021continual,van2022three}.

We expect that the disappointing performance of GEM is due to its choice of objective function: GEM optimizes the loss on the new task (i.e., $\ell_{\text{new}}$) rather than an approximation to the joint loss (i.e., $\widetilde{\ell}_{\text{joint}}$). In other words, we believe that GEM under-utilizes its replay buffer by solely delineating gradient constraints but not actively optimizing the replay-approximated joint loss. When GEM was proposed, it was assumed that directly optimizing a joint loss approximated with a relatively small replay buffer could not work well due to overfitting \citep{lopez2017gradient}, but recent work indicates such overfitting is not as detrimental as thought \citep{chaudhry2019tiny,verwimp2021rehearsal}.
Nevertheless%, despite various extensions of GEM \citep{chaudhry2018efficient,aljundi2019gradient,chen2020revisiting,hu2020gradient}%
, as far as we are aware, changing GEM's objective function has not been explored.

%%%%%%%%%%%%%%%%%%%%%%%%%%%%%%%%%%%%%%%%%%%%

\section{PROOF-OF-CONCEPT EXPERIMENTS}
\label{sec:method}
%In the previous section we discussed that the gradient projection-based optimization routine of GEM ecourages parameter updates that do not strongly interfere with old tasks without imposing overly strict constraints, but that so far this optimization routine has only been used to optimize the loss on the new task. This makes GEM's optimization routine a convenient specimen to ...
As discussed in the last section, the gradient projection-based optimization routine of GEM encourages parameter updates that do not strongly interfere with old tasks without imposing overly strict constraints that would fully segregate tasks. Yet, so far this optimization routine has not been used to optimize proxies of the joint loss. This makes GEM a convenient tool for a first set of proof-of-concept experiments to evaluate the merits of our proposition that continual learning should consider both \emph{what} and \emph{how} to optimize.
%As a first evaluation of the merits of our conceptual proposition, we plan to test whether using GEM's optimization routine to optimize replay-approximated versions of the joint loss can provide the benefits hypothesized in subsection~\ref{sec:alternative}.
We plan to combine GEM's optimization routine both with a basic version of experience replay that explicitly approximates the joint loss (subsection~\ref{sec:poc}) and with state-of-the-art replay-based methods (subsection~\ref{sec:sota}). 

\subsection{Experience Replay with Gradient Projection-based Optimization}
\label{sec:poc}
In a first set of experiments we test whether, when the optimization objective is a standard replay-approximated version of the joint loss, using GEM's gradient projection-based optimization routine provides the benefits hypothesized in subsection~\ref{sec:alternative}.

\paragraph{Approximating the Joint Loss}
\label{sec:joint}
To approximate the joint loss we use a basic version of Experience Replay (ER).
%Particularly, our implementations of ER consider, vanilla experience replay (replay) - storing a small memory of observed data to be interleaved in future training, Dark Experience Replay (DER) \citep{buzzega2020dark} - enhancing vanilla replay by knowledge distillation elements to utilize the predictions of a previous model on the replayed samples, and BiC \citep{wu2019large} - adding a bias correction mechanism for class-incremental learning on top of the LwF \citep{li2017learning} distillation approach. Full details on the mechanisms are provided in the appendix.
In our implementation of ER, at the end of each task new examples are added to the memory buffer using class-balanced sampling from the training set, and in each training iteration uniform sampling from the buffer is used to choose which samples to replay. %Empirically this combination was found to perform strongly in a wide range of scenarios \citep{de2021continual,prabhu2023computationally}. %In the special case of BiC, that requires memory to hold a separate validation set, we split the available buffer resources without increasing them by a training-to-validation ratio of 9:1, as proposed by \citet{wu2019large}.
To approximate the joint loss as closely as possible, when training on the $n$-th task, we balance the loss on the current data and the loss on the replayed data using $\widetilde{\ell}_{\text{joint}} = \frac{1}{n}\ell_\text{{new}} + (1-\frac{1}{n})\widetilde{\ell}_{\text{old}}$. In each iteration, the total number of replayed samples from all past tasks combined is always equal to $b$, which is the size of the mini-batch from the current task.

In addition to approximating the joint loss with ER, we also run experiments using the joint loss itself. For this, all training data from past tasks are stored, and in each iteration we use $b$ samples from each past task to compute $\ell_{\text{old}}$. This can be thought of as `full replay'.

\paragraph{Optimization Trajectory}
To try to improve the sub-optimal optimization trajectory that is taken by standard ER, we use the gradient projection-based optimization routines of GEM and \mbox{A-GEM}. Importantly, we only use the optimization routines of GEM and \mbox{A-GEM}, not their optimization objectives. As optimization objective we instead use the replay-approximated joint loss~$\widetilde{\ell}_{\text{joint}}$. To achieve this, in the description of GEM in the first paragraph of subsection~\ref{sec:gem}, we only need to replace all mentions of $\ell_{\text{new}}$ with $\widetilde{\ell}_{\text{joint}}$. To further illustrate our proposed combination approach, pseudocode for ER\,+\,\mbox{A-GEM} is provided in Algorithm~\ref{algo:ergo}.
%We refer to the resulting approach as \textbf{E}xperience \textbf{R}eplay with \textbf{G}radient projection-based \textbf{O}ptimization (ER+GO).
%The memory and computational costs of ERGO are the same as for GEM.
%ERGO aims to optimize the replay-approximated joint loss with parameter updates that do not strongly interfere with the loss on old tasks.

\begin{algorithm}
    \caption{ER\,+\,\mbox{A-GEM}}\label{algo:ergo}
    \begin{algorithmic}
        \Require{parameters $w$, loss function $\ell$, learning rate $\lambda$, 
        \phantom{for} \phantom{for} \phantom{for} \,\,\, data stream $\{D_1, ..., D_T\}$}
        \vspace{0.04in}
        \State $M \gets \{\}$ %\Comment{Prepare the memory buffer}
        \For{$t=1,...,T$} %\Comment{for every task in the sequence}
            \For {$(x,y) \in  D_t$} %\Comment{for every mini-batch of the current task}
                \State $g \gets \nabla_w \ell(f_w(x), y)$ %\Comment{compute gradient on the current task data}
                \State $(\tilde{x},\tilde{y}) \gets \texttt{SAMPLE}(M)$%\Comment{ sample from the memory buffer}
                \State $g_{\text{old}} \gets \nabla_w \ell(f_w(\tilde{x}), \tilde{y})$
                %\Comment{$\nabla_\theta \ell(f_{\theta}(\tilde{x}), \tilde{y})$} \Comment{compute gradient on the replayed data}
                \State $g_{\text{joint}} \gets \frac{1}{t}g + (1-\frac{1}{t})g_{\text{old}}$ %\Comment{compute gradient of (approximated) joint loss}
                % \If{$g_{\text{old}}^\text{T}g_{\text{joint}}\geq0$}
                %     \State $\bar{g} \gets g_{\text{joint}}$
                % \Else
                %     \State $\bar{g} \gets g_{\text{joint}}-\frac{g_{\text{old}}^{\text{T}}g_{\text{joint}}}{g_{\text{old}}^{\text{T}}g_{\text{old}}}g_{\text{old}}$
                % \EndIf
                \State $\bar{g} \gets \texttt{PROJECT\_AGEM}(g_{\text{joint}}, g_{\text{old}})$ %\Comment{project gradient of joint loss according to projection-operation of GEM or A-GEM}
                %\State $\theta$ ← $\theta - \lambda * \bar{g}$ %\Comment{update the model parameters based on the projected gradient}
                \State $w \gets \texttt{OPTIMIZER\_STEP}(w,\lambda,\bar{g})$
            \EndFor
            \State $M$ ← \texttt{UPDATE\_BUFFER}($M, D_t$) %\Comment{include samples from $D_t$ in the memory buffer}
        \EndFor
        %\vspace{0.01in}
        \algrule
        %\vspace{0.01in}
        \Function{\texttt{PROJECT\_AGEM}}{$g,g_{\text{ref}}$}
            \If{$g^\text{T}g_{\text{ref}}\geq0$}
                \State \Return $g$
            \Else
                \State \Return  $g-\frac{g^{\text{T}}g_{\text{ref}}}{g_{\text{ref}}^{\text{T}}g_{\text{ref}}}g_{\text{ref}}$
            \EndIf
        \EndFunction
    \end{algorithmic}
\end{algorithm}

\paragraph{Approaches to Compare}
The main experimental comparison of interest is between standard ER and our proposed combination approach ER\,+\,GEM, as this allows testing whether, when doing continual learning by optimizing a proxy of the joint loss, benefits can be gained by changing the way this objective is optimized. Additionally, to probe the individual contributions of the optimization objective and the optimization routine, we also include GEM itself and continual finetuning in our experimental comparison. See Table~\ref{tab:compared_methods} for an overview of the approaches we compare. In this table ER can be replaced by `full replay', and GEM can be replaced by \mbox{A-GEM}. We run experiments with all combinations of these base methods.

\begin{table}[!t]
\vskip -0.06in
\caption{\label{tab:compared_methods} Overview of the approaches to compare in our proof-of-concept experiment, illustrated with ER and GEM as base methods. GP: gradient projection.
%Overview of the general comparison for our proof-of-concept experiment. ER: experience replay \{replay, DER, BiC\}; GO: gradient projection optimization \{GEM, A-GEM\}. 
}
\vskip 0.14in
\begin{center}
\begin{tabular}{lP{0.11\textwidth}P{0.11\textwidth}}
    \toprule
     & \bf Approximate & \bf GP-based  \\
    \bf Method & \bf joint loss & \bf optimization \\
    \midrule
    Finetuning & \textcolor{red}{\ding{55}} & \textcolor{red}{\ding{55}}\\
    ER & \textcolor{ForestGreen}{$\pmb{\checkmark}$} & \textcolor{red}{\ding{55}}\\
    GEM & \textcolor{red}{\ding{55}} & \textcolor{ForestGreen}{$\pmb{\checkmark}$}\\
    ER\,+\,GEM & \textcolor{ForestGreen}{$\pmb{\checkmark}$} & \textcolor{ForestGreen}{$\pmb{\checkmark}$}\\
    \bottomrule
\end{tabular}
\end{center}
\end{table}
\vskip 0.24in

\subsection{Improving State-of-the-art}
\label{sec:sota}
Next we ask whether the use of gradient projection-based optimization could improve the performance of state-of-the-art replay-based methods. To test this, we run the methods Dark Experience Replay~(DER; \citealp{buzzega2020dark}) and Bias Correction~(BiC; \citealp{wu2019large}) both with and without using the optimization routine of \mbox{A-GEM}.
%enhancing vanilla replay by knowledge distillation elements to utilize the predictions of a previous model on the replayed samples, and BiC \citep{wu2019large} - adding a bias correction mechanism for class-incremental learning on top of the LwF \citep{li2017learning} distillation approach. In the special case of BiC, that requires memory to hold a separate validation set, we split the available buffer resources without increasing them by a training-to-validation ratio of 9:1, as proposed by \citet{wu2019large}.}
For these experiments we only consider \mbox{A-GEM}, as it is not straight-forward to combine DER with the original version of GEM.
We implement DER and BiC according to their original papers. For completeness, details for both methods are provided in Appendix~\ref{apx:method_details}. As BiC is a method that is specialized for class-incremental learning, it is included only with the class-incremental learning benchmarks.

%%%%%%%%%%%%%%%%%%%%%%%%%%%%%%%%%%%%%%%%%%%%

\section{EXPERIMENTAL PROTOCOL}
\label{sec:experimental_details}

\subsection{Setup}
We consider a task-aware supervised continual learning setting, with a task sequence $\mathcal{T} = \{T_1,...,T_T\}$ of $T$ disjoint classification tasks $T_t$. A fixed capacity neural network model~$f_w$ is incrementally trained on these tasks with a cross-entropy classification loss. When training on task~$T_t$, the model has only access to the training data~$D_t = \{X_t, Y_t\}$ of that task and the data in the memory buffer (see below), and the goal is to learn a model with strong performance on all tasks~$T_{\leq t}$ encountered so far. The model may be evaluated after any parameter update.

\paragraph{Benchmarks} We conduct our study on four benchmarks, covering the domain- and class-incremental learning scenarios \citep{van2022three}.
As class-incremental learning benchmarks we use Split \mbox{CIFAR}-100, which is based on the \mbox{CIFAR}-100 dataset \citep{krizhevsky2009learning}, and Split Mini-Imagenet, which is based on Mini-Imagenet \citep{vinyals2016matching}. Both original datasets contain 50,000 RGB images of 100 classes; \mbox{CIFAR}-100 in resolution 32x32, Mini-Imagenet in resolution 84x84. For Split \mbox{CIFAR}-100 the classes are divided into ten tasks with ten classes each, for Split Mini-Imagenet the classes are divided into twenty tasks with five classes each. In both cases, the classes are divided over the tasks randomly, and for each random seed a different division is used.
We also use the \mbox{CIFAR}-100 dataset to construct \mbox{Domain CIFAR}, a domain-incremental learning benchmark. For this benchmark, each of the twenty super-classes of \mbox{CIFAR}-100 is split across five tasks, such that every task contains one member of each super-class (i.e., there are twenty classes per task, one from each super-class). The goal in each task is to predict to which super-class a sample belongs.
The other domain-incremental learning benchmark is \mbox{Rotated MNIST}. Each task consists of the entire MNIST dataset \citep{lecun1998gradient} with a certain static rotation applied. We construct three tasks with rotations $\{0^\circ, 80^\circ, 160^\circ\}$, as \cite{delange2023continual} found these to provoke the largest stability gaps without inducing ambiguity between digits $6$ and $9$ (which would happen with rotations close to $180^\circ$).

\paragraph{Architectures} For the Rotated MNIST benchmark, we use a fully-connected neural network with two hidden layers of 400 ReLUs each, followed by a softmax output layer. For the other benchmarks, following \cite{lopez2017gradient}, we use a reduced ResNet-18 architecture. Compared to a standard ResNet-18 \citep{he2016deep}, this architecture has three times less channels in each layer and replaces the $7\times7$ kernel with stride of 2 in the initial convolutional layer by a $3\times3$ kernel with stride of 1. The latter prevents an early stark information reduction for images with small resolution.
All benchmarks are trained with a single-headed final layer that is shared between all tasks.

\paragraph{Memory Buffer}
The memory buffer can store up to 100~samples of each class. For the domain-incremental learning benchmarks this means 100 samples of each class per task (e.g., with Rotated MNIST, for each digit the buffer can store 100~examples with rotation~$0^\circ$, 100~examples with rotation~$80^\circ$ and 100~examples with rotation~$160^\circ$). Exceptions to this are the experiments with `full replay', in which all training data are stored.

\paragraph{Offline \& Online}
All experiments are run in both an `offline version' and an `online version'.
In the offline version, multiple passes over the data are allowed, and the number of training iterations is set relatively high to encourage near convergence for each task. In the online version only a single epoch per task is allowed (i.e., each sample is seen just once, with the exception if it is replayed from memory).

\paragraph{Training Hyperparameters}
All models are trained using an SGD optimizer with momentum 0.9 and no weight-decay. When gradient projection is used, this optimizer acts on the projected gradients. Except for whitening (with mean and standard deviation of the respective full training sets), no data augmentations are used. Exceptions to this are the experiments with DER and BiC, for which we use the data augmentations described in the original papers that proposed these methods \citep{buzzega2020dark,wu2019large}. %The data are whitened according to the mean and standard deviation of the respective full training sets, but further no data augmentations are used.
In the offline experiments, we train with mini-batch size 128 for around five epochs (Rotated MNIST) or ten epochs (Domain \mbox{CIFAR}, Split \mbox{CIFAR}-100 and Split Mini-Imagenet) per task. To be exact, we use 2000 iterations per task for Rotated MNIST, 800 for Domain \mbox{CIFAR}, 400 for Split \mbox{CIFAR}-100, and 200 for Split Mini-Imagenet. For each experiment in the offline setting, we sweep a set of static learning rates $\{0.1, 0.01, 0.001 \}$.
For each experiment in the online setting, we sweep both a set of mini-batch sizes $\{10, 64, 128\}$ and a set of static learning rates $\{0.1, 0.01, 0.001\}$. In the online setting, the number of iterations per task is determined by the selected mini-batch size and the number of training samples.

\subsection{Evaluation}
We track the performance of all methods throughout training using `continual evaluation' \citep{delange2023continual}. In particular, after every training iteration we evaluate for each task the accuracy of the model on a hold-out test set.

\paragraph{Testing the Hypotheses}
To quantitatively compare the stability gap of different approaches (i.e.,~to evaluate \textbf{H1}), we use the `average minimum accuracy' metric defined by \cite{delange2023continual}. For completeness, details of this metric are provided in Appendix~\ref{apx:min_acc}. To qualitatively compare the stability gaps, we plot per-task accuracy curves with per-iteration resolution (e.g.,~as in Figure~\ref{fig:stabilit-gap-exemplar-domcif100}).
To compare the learning efficiency of different approaches (i.e.,~to evaluate \textbf{H2}), we use the final average accuracy of the online experiments.
To compare the final learning outcomes of different approaches (i.e., to evaluate \textbf{H3}), we use the final average accuracy of the offline experiments.

\paragraph{Computational Complexity}
To provide insight into the computational complexity of the considered methods, we report for each method its empirical training time on the online version of Split \mbox{CIFAR}-100. For this evaluation all methods are run on identical hardware.

\paragraph{Standard Errors}
Each experiment is run five times, with a different random seed and different division of the classes over tasks for each run. For each metric, both the mean over these runs and the standard error of the mean are reported.

%%%%%%%%%%%%%%%%%%%%%%%%%%%%%%%%%%%%%%%%%%%%

% \section{OUTLOOK}
% % Partly to make clear that experimental results will still be added later! To avoid disappointed readers like our first reviewer.
% Motivated by the stability gap in replay- and regularization-based methods, in this pre-registered report we proposed that strategies for continual learning should focus not only on \emph{what} to optimize, but also on \emph{how}.
% To empirically evaluate the merits of this conceptual proposition, we described a set of proof-of-concept experiments in which gradient projection-based optimization is used to optimize replay-approximated versions of the joint objective.
% We will now perform these experiments by rigorously following the detailed protocol laid out in this report. We expect to communicate the results of these experiments at the \mbox{latest} in~June~2024.

%%%%%%%%%%%%%%%%%%%%%%%%%%%%%%%%%%%%%%%%%%%%

\begin{table*}[!b]
    \centering
    \caption{\textbf{Main quantitative results.} For all benchmarks we report the final average accuracy (AVG) and average minimum accuracy (MIN) for standard ER and incremental joint training (or `full replay') -- both by themselves and in combination with the optimization mechanism of GEM and A-GEM. For each benchmark, this table reports the results obtained with the learning rate (LR) and mini-batch size (BS) that resulted in the highest final accuracy for standard ER and incremental joint training. Bold values mark the highest performance within each `comparison group' (ER, \ergem and \eragem are one such group; Joint, \jointgem and \jointagem another). %; multiple values are marked bold when the metric mean lies withing the interval of the standard error. 
    Reported is the mean $\pm$ standard error over five random seeds. The full quantitative results for all pre-registered experiments are provided in Tables~\ref{apx_table:rotmnist}-\ref{apx_table:miniimg} in Appendix~\ref{apx:additional_results}.
    }
    \vspace{1em}
    %\resizebox{0.48\textwidth}{!}{
\newcommand{\HiER}[1]{\pmb{#1}}
\newcommand{\HJ}[1]{\pmb{#1}}
    \begin{tabular}{p{5em} p{3em} p{2em} | p{3.8em}p{3.8em}p{3.8em} | p{3.8em}p{3.8em}p{3.8em}p{0em}}
        & & & \centering \vspace{-0.25em} ER & \centering ER + GEM & \centering ER + A-GEM & \centering \vspace{-0.25em} Joint & \centering Joint + GEM & \centering Joint + A-GEM &\\
        \midrule
        \multirow{4}{5em}{Rotated MNIST} & \multirow{2}{3em}{Offline \\ \hspace*{0.5em}\tiny\mbox{LR 0.1}} 
        & MIN 
        & $83.1 \pm 0.5$ & $\HiER{84.1 \pm 0.4}$ & $82.5 \pm 0.7$ 
        & $\HJ{86.7 \pm 0.8}$ & $\HJ{87.5 \pm 0.9}$ & $\HJ{86.6 \pm 0.6}$ \\
        & 
        & AVG & $91.9 \pm 0.1$ & $\HiER{93.7 \pm 0.1}$ & $91.8 \pm 0.2$ 
        & $97.5 \pm 0.0$ & $\HiER{97.8 \pm 0.0}$ & $97.5 \pm 0.0$\\
        \cmidrule{3-9}
        & \multirow{2}{5em}{Online \\ {\tiny\hspace*{-1em}\mbox{LR 0.01; BS 10}}} 
        & MIN 
        & $86.8 \pm 0.3$ & $\HiER{89.1 \pm 0.4}$ & $87.1 \pm 0.4$ 
        & $92.3 \pm 0.4$ & $\HJ{92.8 \pm 0.3}$ & $92.4 \pm 0.1$\\
        & 
        & AVG & $92.7 \pm 0.1$ & $\HiER{94.2 \pm 0.1}$ & $92.8 \pm 0.2$
        & \HJ{$96.8 \pm 0.0$} & $\HJ{96.8 \pm 0.1}$ & \HJ{$96.8 \pm 0.1$}\\
        \midrule
        \multirow{4}{5em}{Domain \mbox{CIFAR-100}} & \multirow{2}{3em}{Offline\\ \hspace*{0.5em}\tiny\mbox{LR 0.1}} 
        & MIN 
        & $\HiER{33.2 \pm 1.3}$ & $~~7.2 \pm 2.6$ & $\HiER{34.0 \pm 0.9}$ 
        & $\HJ{42.9 \pm 0.7}$ & $\HJ{42.2 \pm 1.1}$ & $\HJ{42.7 \pm 0.8}$\\
        & 
        & AVG & $\HiER{48.6 \pm 0.5}$ & $23.9 \pm 1.8$ & $\HiER{48.6 \pm 0.5}$
        & $\HJ{52.4 \pm 0.8}$ & $\HJ{52.6 \pm 0.6}$ & $\HJ{52.0 \pm 0.7}$\\
        \cmidrule{3-9}
        & \multirow{2}{5em}{Online\\ {\tiny\hspace*{-1em}\mbox{LR 0.01; BS 10}}} 
        & MIN & $\HiER{29.2 \pm 0.7}$ & $~~4.3 \pm 0.1$ & $\HiER{29.2 \pm 0.7}$ 
        & $\HJ{35.5 \pm 1.1}$ & $\HJ{35.6 \pm 1.3}$ & $\HJ{35.1 \pm 0.9}$\\
        & 
        & AVG & $\HiER{38.3 \pm 0.8}$ & $19.8 \pm 1.4$ & $\HiER{38.3 \pm 0.8}$ 
        & $\HJ{49.8 \pm 1.0}$ & $\HJ{49.4 \pm 1.5}$ & $\HJ{49.7 \pm 1.2}$\\
        \midrule
        \multirow{4}{5em}{Split \mbox{CIFAR-100}} & \multirow{2}{3em}{Offline \\ \hspace*{0.5em}\tiny\mbox{LR 0.1}} 
        & MIN & $\HiER{12.4 \pm 0.3}$ & $~~0.0 \pm 0.0$ & $\HiER{12.1 \pm 0.3}$ 
        & $22.5 \pm 0.2$ & $~~2.8 \pm 1.9$ & $\HJ{23.1 \pm 0.4}$\\
        & 
        & AVG & $\HiER{22.8 \pm 0.4}$ & $~~7.1 \pm 1.9$ & $\HiER{22.4 \pm 0.6}$ 
        & $\HJ{32.2 \pm 0.5}$ & $16.5 \pm 6.7$ & $\HJ{32.5 \pm 0.4}$\\
        \cmidrule{3-9}
        & \multirow{2}{5em}{Online\\ {\tiny\hspace*{-1em}\mbox{LR 0.01; BS 10}}} 
        & MIN & $\HiER{10.6 \pm 0.3}$ & $~~0.0 \pm 0.0$ & $\HiER{10.7 \pm 0.4}$ 
        & $\HJ{27.0 \pm 0.5}$ & $\HJ{27.2 \pm 0.5}$ & $\HJ{27.2 \pm 0.6}$\\
        & 
        & AVG & $\HiER{20.7 \pm 0.3}$ & $~~8.5 \pm 0.9$ & $\HiER{20.5 \pm 0.4}$ 
        & $\HJ{41.2 \pm 0.6}$ & $\HJ{41.1 \pm 0.4}$ & $\HJ{41.4 \pm 0.2}$\\
        \midrule
        \multirow{4}{3em}{\mbox{Split Mini-} \mbox{ImageNet}} & \multirow{2}{3em}{Offline\\ \hspace*{0.5em}\tiny\mbox{LR 0.1}} 
        & MIN & $\HiER{~~8.6 \pm 0.5}$ & $~~0.0 \pm 0.0$ & $\HiER{~~8.3 \pm 0.4}$
        & $\HJ{16.5 \pm 0.5}$ & $~~0.0 \pm 0.0$ & $\HJ{16.4 \pm 0.4}$\\
        & 
        & AVG & $\HiER{16.9 \pm 0.9}$ & $~~4.7 \pm 1.6$ & $\HiER{16.9 \pm 0.5}$ 
        & $\HJ{28.1 \pm 0.6}$ & $~~3.3 \pm 0.4$ & $\HJ{28.5 \pm 0.7}$\\
        \cmidrule{3-9}
        & \multirow{2}{3em}{Online\\ {\tiny\hspace*{-1em}\mbox{LR 0.01; BS 64}}} 
        & MIN 
        & $\HiER{~~3.3 \pm 0.2}$ & $~~0.0 \pm 0.0$ & $\HiER{~~3.2 \pm 0.2}$ 
        & $\HJ{~~9.1 \pm 0.2}$ & $~~0.0 \pm 0.0$ & $\HJ{~~8.9 \pm 0.3}$\\
        & 
        & AVG & $\HiER{13.9 \pm 0.5}$ & $~~5.8 \pm 0.3$ & $\HiER{14.6 \pm 0.3}$ 
        & $\HJ{27.9 \pm 0.8}$ & $~~5.5 \pm 0.5$ & $\HJ{28.3 \pm 1.3}$\\ 
    \end{tabular}
    %}
    \label{tab:results}
\end{table*}

\section{RESULTS}
\label{sec:results}
% Summerization of the main results
In Table~\ref{tab:results} we report the quantitative results for our main experimental comparisons across all benchmarks. The average minimum accuracy (MIN) indicates the worst-case accuracy throughout training, which we use as proxy for the stability gap. The final average accuracy (ACC) reflects the performance of the continual learning model at the end of training.
% case measure to indicate the impact of the stability gap, and by the
%which shows the final average accuracy (AVG) and average minimum accuracy (MIN) across all benchmarks. It contrasts the performance of continual learning mechanisms with a worst-case measure that is determined by the stability gap.
%We find that \eragem has barely any impact on performance when compared to standard ER, while \ergem shows an effect that is, however, not always beneficial.
The results indicate that combining standard~ER, or incremental joint training, with the optimization mechanism of A-GEM does not significantly change either the stability gap or the final performance on any of the tested benchmarks.
%For Rotated MNIST, \ergem consistently narrows the stability gap and enhances overall performance. This improvement can also be observed for incremental joint training. Although the effect is modest, it suggests that, in principle, gradient projection can minimize the stability gap (H1) and, by extension, improve final performance (H3). Moreover, this demonstrates the benefit of adopting a more careful optimization approach.
On the other hand, combining ER or incremental joint training with GEM's optimization mechanism induces clear effects, although these are not always beneficial. 
For Rotated MNIST, using the optimization mechanism of GEM on top of ER reduces the stability gap, as reflected by an increase of the average minimum accuracy, and improves final performance. These positive effects of GEM's optimization mechanism persist even when used on top of incremental joint training (see also Table~\ref{apx_table:rotmnist} in the Appendix).
However, for the other benchmarks, using the optimization mechanism of GEM does not yield these benefits and instead frequently impairs performance.
%not confirm our hypotheses. For these more complex benchmarks, \ergem does not enhance accuracy, but often disrupts the performance observed with standard ER. 
%Also, for these benchmarks, since the stability gap does not decrease for either offline or online learning, we cannot achieve the possible benefits of improving final performance (H3) or learning efficiency (H2) by closing the gap. Therefore, we cannot test these hypotheses for CIFAR-100 and Mini-Imagenet datasets.

In the following, we first examine the stability gap for a selection of benchmarks in more detail~(subsection~\ref{subsec:results_h1}-\ref{subsec:online}). Then, we look at the results for extending recent state-of-the-art replay-based methods with A-GEM's optimization mechanism (subsection~\ref{subsec:der_and_bic}). Finally, we evaluate the additional computational overhead induced by the GEM and A-GEM gradient projection mechanisms (subsection~\ref{subsec:comp}).
%In the following, we present more detailed examination, including per-iteration evaluation of the stability gap for a selection of benchmarks.  A comprehensive tabular report of our results across all experiments, is provided in Appendix\ref{apx:additional_results}.

\begin{figure*}[!b]
    \centering
    \includegraphics[width=0.99\textwidth]{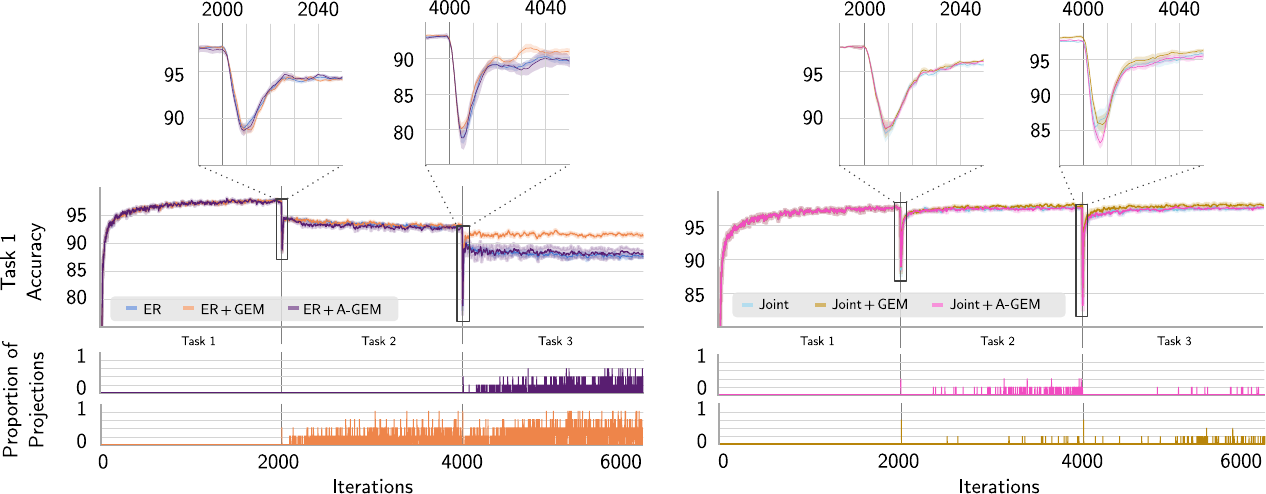}
    \vspace{0.02in}
    %\caption{\textbf{Stability gaps for Task 1 of offline learning incremental Rotated MNIST.} The left side shows standard ER, the right side Incremental Joint training -- both in combination with orthogonal gradient projection by GEM and A-GEM. The top section shows a zoomed view of the stability gap, while the bottom section indicates the proportion of runs where the gradient was projected. A value of $0.0$ signifies that in no run a gradient was projected at this iteration, conversely $1.0$ indicates that all runs projected a gradient. Each setting comprises $5$ runs with different random seeds and uses a learning rate of $0.1$. The plots show mean $\pm$ standard error as the shaded area, aggregated over all runs.}
    \caption{\textbf{Stability gaps for the first task of offline Rotated MNIST.} The left side shows standard ER, the right side incremental joint training (or `full replay') -- both by themselves and in combination with the optimization mechanism of GEM and A-GEM. The middle panels show the test accuracy on the first task while the model is incrementally trained for all tasks of the benchmark. The top panels show zoomed-in views of the first 50 training iterations after a task switch, allowing a more detailed qualitative comparison of the stability gap. These plots show the mean $\pm$ standard error (shaded area) over five runs with different random seeds. The bottom panel shows for every iteration the proportion of runs where the gradient was projected, with $0$ indicating that at this iteration there was no run in which a gradient was projected and~$1$~indicating that there was a gradient projection in every run.}
    \label{fig:results_rot_mnist_off}
\end{figure*}

\subsection{Rotated MNIST}
\label{subsec:results_h1}
% Offline Part
%As outlined in the previous section, domain-incremental Rotated MNIST provides a case that demonstrates the merit of altering the optimization trajectory by gradient projection. From Table~\ref{tab:results} we already noted an increase in average minimum accuracy, i.e. a reduction of stability gap on average. This is also visible in the detailed views of the stability gap in the upper part of Figure~\ref{fig:results_rot_mnist_off}.
%From this figure we also recognize that it is the additional gradient projection that brings the positive effect on the stability gap. The bottom part of the figure marks the percentile of runs that did gradient projections, per iteration. Most runs exhibit gradient projections right after a task switch, even if there are barely any projections otherwise. The case of \jointgem is a prominent example of this. At the times where there are no projections, the addition of +GEM or +\,A-GEM has no effect on the gradient, i.e. \jointgem reduces to standard joint training. This means it is the projections that happen after the task switch - which is at the same time when the stability gap is opening - that improve the performance. 
%We note that the same correlation is visible for \jointagem as well, but without substantial impact on the final performance. 
Our results in Table~\ref{tab:results} hint that the domain-incremental learning problem of Rotated MNIST provides a case where replay can be improved by altering the optimization trajectory with GEM's gradient projection mechanism. 
%From Table~\ref{tab:results} we note an increase in average minimum accuracy, i.e. on average a reduction of the stability gap. 
In Figure~\ref{fig:results_rot_mnist_off}, we depict the per-iteration accuracy on the first task while incrementally training on all the tasks, with detailed views at the task switches to highlight the stability gap. We observe that when training on the third task, \ergem modestly reduces the stability gap compared to ER. Regarding the final performance, \ergem shows a clear improvement relative to ER. 
%Regarding the stability gap, the detailed views of the stability gap in the upper part of Figure~\ref{fig:results_rot_mnist_off} show that its decrease is not consistent for every newly learned task. 
%This is also visible in the detailed views of the stability gap in the upper part of Figure~\ref{fig:results_rot_mnist_off}. Here, \ergem modestly reduces the stability gap on the first task when learning the third task, and clearly improves the average final performance compared to standard ER. 
For incremental joint training, the effects of using GEM's optimization routine are more modest, which might be related to the relatively low number of gradients being projected, as indicated in the bottom part of the figure. The stability gap is almost unaltered, but we still observe a statistically significant improvement of the average final performance. 
%which can be partially explained by the reduced number of gradients being projected illustrated in the bottom part of the figure. Yet still, we find a statistically significant improvement of the average final performance. 
In contrast, using \mbox{A-GEM}'s optimization routine does not yield any improvements on top of ER or incremental joint training.
%We note that in terms of the number of projections \eragem is comparable to \ergem, however, without any benefits over standard ER in terms of performance or addressing the stability gap. 
In Appendix~\ref{sec:gem_memory_strength}, we investigate a hyperparameter of GEM that can explain part of this difference between \ergem and \eragem.

% Domain CIFAR-100
\begin{figure*}[!t]
    \centering
    \vspace{1em}
    \includegraphics[width=0.99\textwidth]{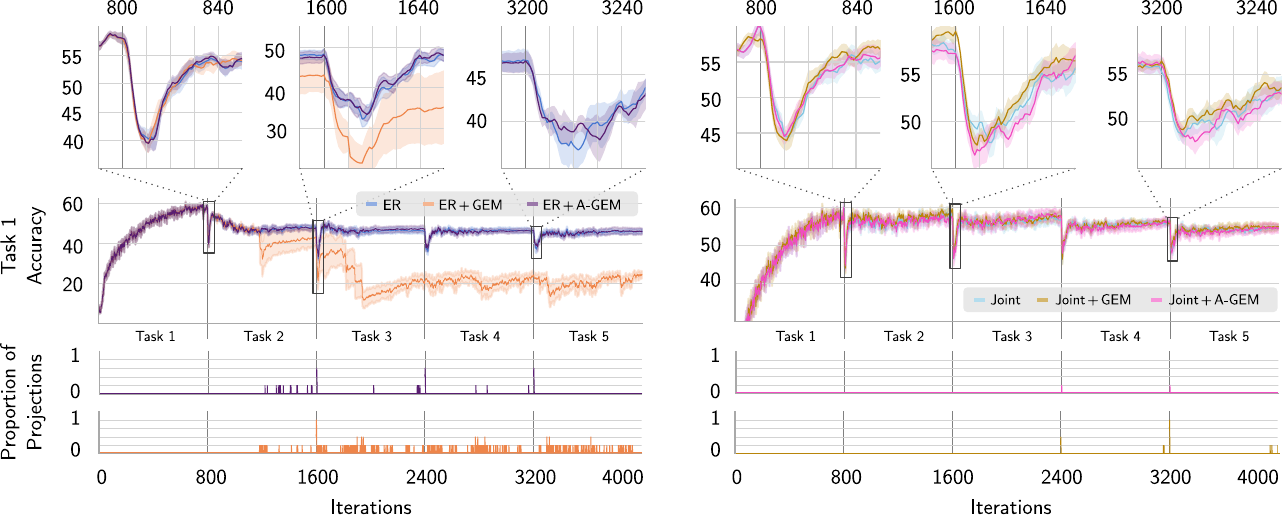}
    \vspace{0.02in}
    \caption{\textbf{Stability gaps for the first task of offline  Domain CIFAR-100.} The left side shows standard ER, the right side incremental joint training (or `full replay') -- both by themselves and in combination with the optimization mechanism of GEM and A-GEM. The middle panels show the test accuracy on the first task while the model is incrementally trained for all tasks of the benchmark. The top panels show zoomed-in views of the first 50 training iterations after a task switch, allowing a more detailed qualitative comparison of the stability gap. These plots show the mean $\pm$ standard error (shaded area) over five runs with different random seeds. The bottom panel shows for every iteration the proportion of runs where the gradient was projected, with $0$ indicating that at this iteration there was no run in which a gradient was projected and~$1$~indicating that there was a gradient projection in every run.}
    \label{fig:results_domcif_off}
    \vspace{0.4em}
\end{figure*}

% Split-CIFAR-100
\begin{figure*}[!b]
    \centering
    \includegraphics[width=0.99\textwidth]{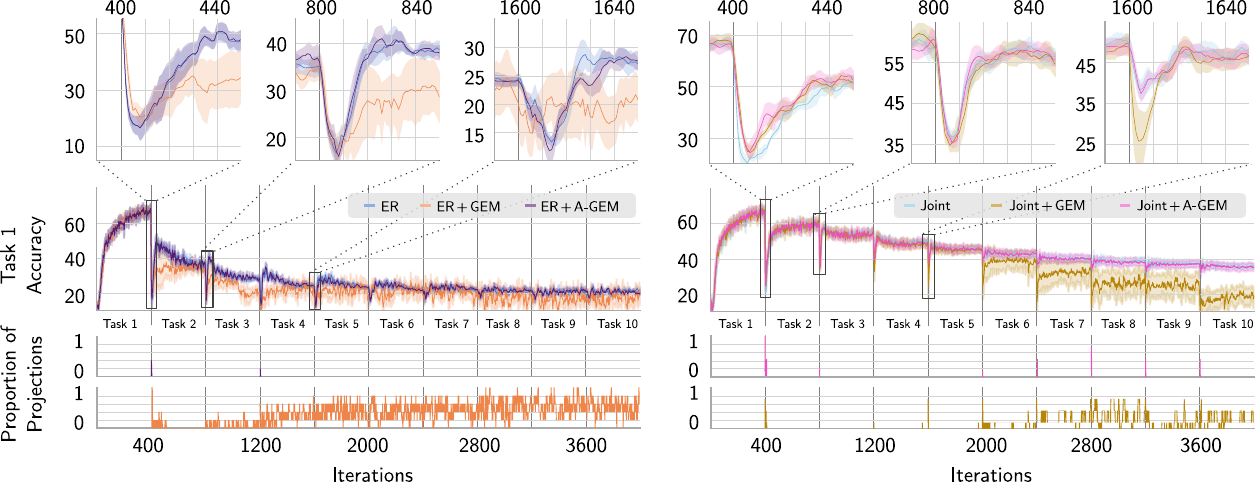}
    \vspace{0.02in}
    \caption{\textbf{Stability gaps for the first task of offline  Split CIFAR-100.} The left side shows standard ER, the right side incremental joint training (or `full replay') -- both by themselves and in combination with the optimization mechanism of GEM and A-GEM. The middle panels show the test accuracy on the first task while the model is incrementally trained for all tasks of the benchmark. The top panels show zoomed-in views of the first 50 training iterations after a task switch, allowing a more detailed qualitative comparison of the stability gap. These plots show the mean $\pm$ standard error (shaded area) over five runs with different random seeds. The bottom panel shows for every iteration the proportion of runs where the gradient was projected, with $0$ indicating that at this iteration there was no run in which a gradient was projected and~$1$~indicating that there was a gradient projection in every run.}
    \label{fig:results_cif100_off}
    \vspace{1em}
\end{figure*}

\subsection{Offline Natural Image Benchmarks}
In the offline versions of the continual learning benchmarks based on CIFAR-100 and Mini-Imagenet, the benefits we observed with \ergem on Rotated MNIST do not recreate.
%The beneficial effects we previously observed on Rotated MNIST do not recreate for the offline training setting for continual learning benchmarks based on CIFAR-100 and Mini-Imagenet.
Instead, \ergem substantially deteriorates the performance on these benchmarks. As before, using \mbox{A-GEM}'s optimization routine does not seem to have any real impact.  % but also average performance. 

Detailed views for Domain CIFAR-100 are depicted in Figure~\ref{fig:results_domcif_off}. The performance of \ergem decreases incrementally due to a collapse originating from the combination of ER with GEM's gradient projection.
%We analyze this behavior further in Section~\ref{sec:gem_memory_strength}. 
For Joint\,+\,GEM, a collapse does not arise but there is also no benefit over Joint, which may be related to the low number of gradient projections.
For Split CIFAR-100, detailed in Figure~\ref{fig:results_cif100_off}, we find slightly varied behavior.
%The class-incremental setting, introduces larger distribution shifts and a larger number of tasks. 
With ER\,+\,GEM, there is a small reduction of the stability gap after the first task switch, however, seemingly at the expense of model's ability to recover lost performance.
For Joint\,+\,GEM, collapse is visible from task~6 onwards, but already during earlier task switches, such as when starting training on task~5, there is an increased instability that is enlarging the stability gap instead of closing it. 
The results for Split Mini-ImageNet, provided in Figure~\ref{fig:miniimgnet_offline} in Appendix~\ref{apx:additional_results}, are comparable to those of Split CIFAR-100.

% Domain CIFAR-100 Online
\begin{figure*}[!t]
    \centering
    \includegraphics[width=0.99\textwidth]{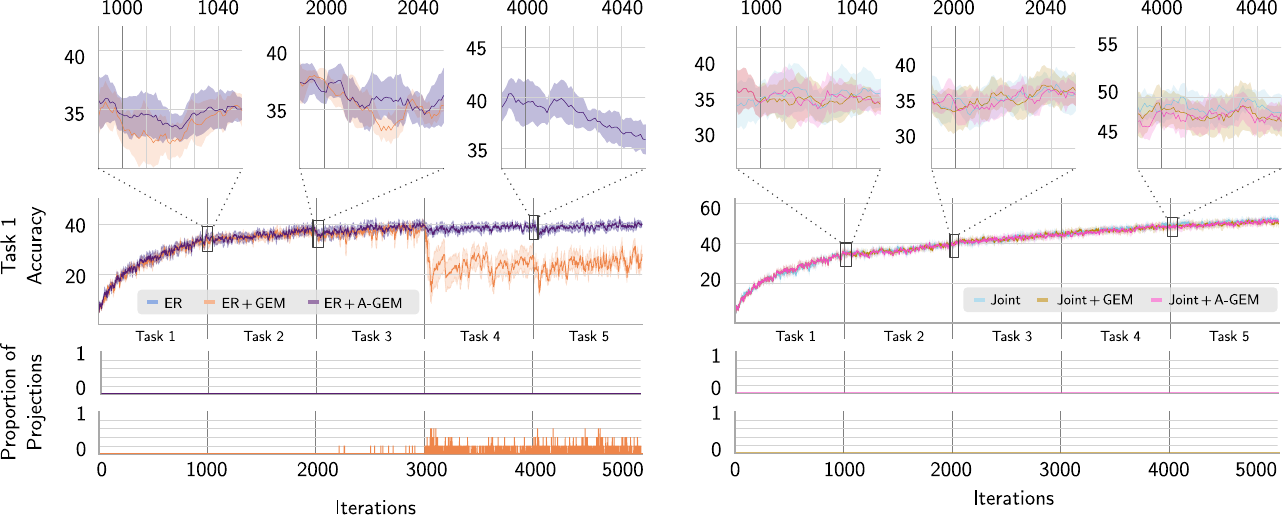}
    \vspace{0.005in}
    \caption{\textbf{Stability gaps for first task of online Domain CIFAR-100.} The left side shows standard ER, the right side incremental joint training (or `full replay') -- both by themselves and in combination with the optimization mechanism of GEM and A-GEM. The middle panels show the test accuracy on the first task while the model is incrementally trained for all tasks of the benchmark. The top panels show zoomed-in views of the first 50 training iterations after a task switch, allowing a more detailed qualitative comparison of the stability gap. These plots show the mean $\pm$ standard error (shaded area) over five runs with different random seeds. The bottom panel shows for every iteration the proportion of runs where the gradient was projected, with $0$ indicating that at this iteration there was no run in which a gradient was projected and~$1$~indicating that there was a gradient projection in every run.}
    \label{fig:results_domcif_online}
    \vspace{-0.26em}
\end{figure*}

%%%%%%%%%%%%%%%%%%%%%%%% 
% Table for DER and BIC
\begin{table*}[!b]
    \centering
    \vspace{-0.26em}
    \caption{\textbf{Using A-GEM on top of state-of-the-art methods.} For the benchmarks in the offline setting, we report the final average accuracy (AVG) and average minimum accuracy (MIN) for DER and BiC -- both by themselves and in combination with A-GEM's optimization mechanism. This table shows the results for the learning rate (LR) with the best final average accuracy for the standard versions of DER and BiC. Reported is the mean $\pm$ standard error over five random seeds.}
    \vspace{0.88em}
    %\resizebox{0.48\textwidth}{!}{
    \begin{tabular}{p{5em} p{1.5em} p{2em} | p{4em}p{4em}|p{4em}p{4em}p{0em}}
        & \centering \vspace*{-0.25em} &  
        & \centering \vspace*{-0.25em} DER & \centering DER + A-GEM & \centering \vspace*{-0.25em} BiC & \centering BiC + A-GEM & \\
        %\cmidrule{2-6}
        \midrule
        \multirow{2}{5em}{Rotated MNIST}
        & \multirow{2}{5em}{\hspace*{-0.5em}\scriptsize\mbox{LR $0.01$}} 
        & MIN & $82.6 \pm 0.5$ & $83.0 \pm 0.4$ & \centering - & \centering - &\\
        & & AVG & $87.3 \pm 0.2$ & $87.1 \pm 0.2$ & \centering - & \centering - &\\ 
        \midrule
        \multirow{2}{5em}{Domain \mbox{CIFAR-100}}
        & \multirow{2}{5em}{\hspace*{-0.5em}\scriptsize\mbox{LR $0.01$}}
        & MIN & $44.4 \pm 0.7$ & $44.8 \pm 1.0$ & \centering - & \centering - &\\
        & & AVG & $59.7 \pm 0.5$ & $60.7 \pm 0.4$ & \centering - & \centering - &\\
        \midrule
        \multirow{2}{5em}{Split \mbox{CIFAR-100}}  
        & \multirow{2}{5em}{\hspace*{-0.5em}\scriptsize\mbox{~LR $0.1$}} 
        & MIN & $~~1.4 \pm 0.1$ & $~~1.6 \pm 0.2$ & $22.3 \pm 0.9$ & $23.1 \pm 1.1$ \\
        & & AVG & $26.0 \pm 0.9$ & $25.6 \pm 1.3$ & $38.3 \pm 1.0$ & $38.4 \pm 0.9$\\
        \midrule
        \multirow{2}{5em}{Split Mini- \mbox{ImageNet}}  
        & \multirow{2}{5em}{\hspace*{-0.5em}\scriptsize\mbox{LR $0.01$}}
        & MIN & $~~0.1 \pm 0.0$ & $~~0.2 \pm 0.1$ & $13.3 \pm 0.3$ & $13.5 \pm 0.2$\\
        & & AVG & $~~7.4 \pm 0.5$ & $~~7.6 \pm 0.7$ & $25.9 \pm 0.7$ & $26.2 \pm 0.5$\\
    \end{tabular}
%    }
    \label{tab:results_der_bic}
    \vspace{-0.25em}
\end{table*}
%%%%%%%%%%%%%%%%%%%%%%%%%%%

\subsection{Online Benchmarks}
\label{subsec:online}
% Online Part
%This implies that the gradients obtained from the replayed data align more strongly with the gradients from the current task’s batch, as they optimize both old and new task. Therefore, not only is the stability gap generally reduced, but also, the gradient projection becomes ineffective. Figure~\ref{fig:results_domcif_online} shows an example of this. For \eragem there is not a single gradient projection, meaning it fully reduced to standard ER. For \ergem there are projections during later tasks, which are but mostly disrupting to the learning as can be observed from the increased noise and performance collapse.\\
After running the experiments that we had proposed for the online versions, we found them to be less insightful than we had anticipated.
On the one hand, for Rotated MNIST, due to its relatively large training set and rapid convergence of the model, the results for the online version are largely similar to those for the offline version, and therefore do not provide additional insights.
On the other hand, for the natural image benchmarks, we found that for the settings that we had chosen for the online versions, there are no clear stability gaps with standard~ER or with incremental joint training. This is illustrated for Domain CIFAR-100 in Figure~\ref{fig:results_domcif_online}. It can further be seen that there are very few gradient projections, which might be related to the absence of a stability gap in these training conditions.
%Contrary to our expectations when stating the experimental protocol, the online continual learning setting proved less interesting to our case. %The ineffectiveness of gradient projection from the GEM and A-GEM mechanism prevent meaningful insights beyond judging the particular suitability of these mechanisms for improving the optimization trajectory.
%For all benchmarks except Rotated MNIST, the number of iterations per task is reduced to the extent that the model does not fully converge to a locally optimal solution. One finds that using standard ER improves accuracy on previous tasks in continued learning. In turn, the stability gaps are generally less severe compared to the offline training counterpart, and furthermore, the gradient projection of GEM and A-GEM becomes ineffective. This can be observed from the bottom of Figure~\ref{fig:results_domcif_online}. It exemplifies the effects listed above for the incremental Domain CIFAR-100 benchmark and, at the bottom of the figure, shows that barely any projections are happening. 
For \eragem there is not a single projection, meaning it fully reduces to standard ER, while
for \ergem there are projections during the training of later tasks, which seem mostly disruptive to the learning as indicated by the increased noise and subsequent performance collapse.
We find similar results for the other natural image benchmarks. For Split CIFAR-100, shown in Figure~\ref{fig:cif100_online} in the Appendix, there is only a small stability gap and a few A-GEM projections after the first task switch, but after that not anymore.

\subsection{Combining DER and BiC with A-GEM}
\label{subsec:der_and_bic}
In Table~\ref{tab:results_der_bic}, we report the results for combining the recent state-of-the-art replay-based methods DER and BiC with the gradient projection-based optimization mechanism of A-GEM. Similar as in our main experiments in Table~\ref{tab:results}, we find no consistent improvements by using \mbox{A-GEM}'s optimization mechanism on top of either DER or BiC. 
%We note that on Rotated MNIST, perhaps surprisingly, DER performs worse than standard ER. We found this to be explained by DER's addition of data augmentations that are not useful for Rotated MNIST. 

\subsection{Computational Complexity}
\label{subsec:comp}

\begin{table}[!t]
    \centering
    \caption{\textbf{Computational complexity.} The empirical training time on online Split CIFAR-100 is shown. Reported is the mean $\pm$ standard error over five runs with learning rate~$0.01$ and mini-batch size~$10$.}
    \vspace{1em}
    %\resizebox{0.3\textwidth}{!}{
    % \begin{tabular}{p{4.5em} p{1.2em} p{2em} | p{4em}p{4em}p{4em}p{4em}p{0em}}
    %     & \centering \vspace*{-0.25em} LR & Final ACC 
    %     & \centering \vspace*{-0.25em} DER & \centering DER +GEM & \centering \vspace*{-0.25em} BiC & \centering BiC +AGEM & \\
    %     %\cmidrule{2-6}
    %     \midrule
    %     \multirow{2}{5em}{Rotated MNIST}
    %     & \multirow{2}{5em}{$0.01$} 
    %     & MIN & $82.6 \pm 0.5$ & $83.0 \pm 0.4$ & \centering - & \centering - &\\
    %     & & AVG & $87.3 \pm 0.2$ & $87.1 \pm 0.2$ & \centering - & 
    % \end{tabular}

    \begin{tabular}{p{6em} | p{5.5em} p{0em}}
         & \centering Time [s] & \\
         \midrule
         %FT & $~~147.3 \pm 1.8$\\
         %GEM & $1119.3 \pm 22.8$ \\
         %A-GEM & $~~266.9 \pm 3.5$\\[0.5em]
         %
         ER & $~~59.24 \pm 0.66$\\
         ER\,\mbox{+\,A-GEM} & $~~61.52 \pm 0.60$\\
         ER\,+\,GEM & $252.25 \pm 1.21$ \\[0.5em]
         %
         %DER & $~~149.7 \pm 0.8$\\
         %DER \mbox{+ A-GEM} & $~~480.4 \pm 0.9$\\[0.5em]
         %
         %BiC & $~~200.6 \pm 7.1$\\
         %BiC \mbox{+ A-GEM} & $~~359.1 \pm 5.4$ 
    \end{tabular}
%    }
    \label{tab:timing}
\end{table}

In Table~\ref{tab:timing}, we evaluate the computational complexity of \ergem and \eragem relative to standard ER, by comparing their empirical training time on the online version of Split CIFAR-100. For a fair comparison, all methods are run on the same hardware and we attempted to use comparable implementations.
We observe that \ergem takes the longest time to train. This can be explained because \ergem needs to compute a reference gradient for each previous task and solve the quadratic program integrated in its gradient projection mechanism.
The difference in training time between \eragem and standard ER is relatively small.
%but still more than standard ER, again, because of the one additional backward pass that is required to calculate its reference gradients.\\
%In general, the computational complexities of the methods used in this work are largely governed by the number of forward and backward passes.
The overhead introduced by \eragem compared to standard ER is governed by the line~$\bar{g} \gets \texttt{PROJECT\_AGEM}(g_{\text{joint}}, g_{\text{old}})$ in Algorithm~\ref{algo:ergo}: the only additional computations are the dot-product~$g^{\text{T}}g_{\text{ref}}$ and the gradient projection~$g - \frac{g^{\text{T}}g_{\text{ref}}}{g^{\text{T}}_{\text{ref}}g_{\text{ref}}}$. Extra computations to calculate the reference gradients, which are required for \mbox{A-GEM}'s projection, are not needed because those are already available from the ER mechanism.

%%%%%%%%%%%%%%%%%%%%%%%%%%%%%%%%%%%%%%%%%%%%

\section{DISCUSSION}
\label{sec:discussion}
% summarize the conceptual argument (section 2) almost similar to abstract (even with perfect approixmation of joint loss you find stability gap / most methods optimize loss -> stability gap still there -> need something else / this leads to complementary (useful) perspectives which could be a good way to think about developing new continual learning methods)
% descripe pre-registered proof of concept experimetns (use existing methods that change optimization trajectory and combine with replay / a specific implementation + explain we did not find clear and consistent effects / )
% thoughts on pre-registered format
    % our hypothesis were too open - succes of experiments provides support but not successful experiments could not disprove the hypothesis - caveat of our hypothesis
    % the importance of being precises - "we said to use the gem mechanism" which was ambigous because 
    % checking the experimental setup before - we should have run the expereince replay benchmarks in advance (ideally we should have tested this before - however extending to the acual experiments would be relatively easy which requires increased trust for the pre-registed protocol)
% the conceptual proposal "still holds" - future work on how to test it next / 

Most of the recent progress in continual learning has been achieved by the addition of replay or regularization terms to the loss function, which is done typically with the aim of approximating the loss that one would like to optimize (e.g.,~the joint loss over all tasks so far). However, in this work we have shown that even if current replay- or regularization-based methods would manage to perfectly approximate the desired loss, they would still suffer from the stability gap. This insight highlights that the current approach to continual learning is not sufficient. This has motivated us to argue that rather than concentrating only on improving the optimization objective (i.e.,~what loss function to optimize), continual learning should also focus on improving the optimization trajectory (i.e.,~how to optimize that loss function). Our proposition thus delineates two complementary perspectives to continual learning, which we believe to be a useful framework for developing new continual learning methods.

In search of a proof-of-concept demonstration for our proposition, we proposed a series of pre-registered experiments combining established continual learning methods that approximate the joint loss by experience replay, with existing continual learning methods that change the optimization trajectory. These existing optimization-based continual learning methods, namely GEM and A-GEM, had previously only been used to optimize the loss on the new task, and not -- as we propose -- to optimize an approximated version of the joint loss.
Unfortunately, these pre-registered experiments did not show clear and consistent benefits of this combined approach. In particular, altering the optimization trajectory with A-GEM did not show any substantial impact on the results compared to standard replay. On the other hand, using GEM’s optimization on top of standard replay had positive effects on the Rotated MNIST benchmark, but also a disruptive effect on the performance on other benchmarks composed of natural images. 

%Looking back at the pre-registered format we became aware of caveats that went unnoticed in the initial proposition phase. We realized that our hypotheses are formulated in such a way that they could be confirmed by our pre-registered experiments, but not rejected. Our inconclusive findings for \ergem do not invalidate our conceptual proposal. Regarding the implementation of GEM’s optimization mechanism we found the impact of its hyper-parameter $\gamma$ to be higher than expected and we added an entire set of experiments to gain further insights. Similarly, the influence of batch-normalization to the experiments became an overlooked detail with some importance to the final implementation. Regarding the online learning experiments, in some settings the model did not show a stability gap anymore. In hindsight it would have been advantageous to run preliminary baseline tests and maybe change the setting. Yet such preliminary experiments the extension to the full set of pre-registered experiments would have been straightforward. This raises the required trust that the un-conference process is actually followed by the authors.

While our pre-registered experiments were not able to confirm our hypotheses or provide convincing proof-of-concept demonstrations, we emphasize that the contributions of this paper extend beyond those empirical results. Indeed, taking inspiration from the stability gap \citep{delange2023continual}, the main contribution of this paper has been to develop the conceptual proposition that continual learning should ask not only \emph{what} to optimize, but also \emph{how}. Following the release of the pre-registered proposal of this paper, our proposition has started to inspire follow-up work. Building on our work, \citet{kamath2024expanding} demonstrate that the stability gap can occur even with incremental learning of the same task (i.e.,~when there is no distribution shift). In other recent work, \citet{yoo2024layerwise} provide supporting evidence for our proposition by showing that in an online continual learning setup, the performance of various established replay methods can be improved by using a proximal point method-inspired optimization routine.

Although we have provided compelling arguments that the way in which optimization is done in continual learning can be improved, how to do this largely remains an open question that demands further exploration. 
Promising directions of future work include the development of more principled methods for gradient projection to bridge the gap between theoretical guarantees and practical performance. Beyond gradient projection, other constrained optimization mechanisms could be explored, such as using Lagrange multipliers (e.g.,~building upon \citealp{elenter2023primal}) or trust region methods (e.g.,~building upon \citealp{kao2021natural}). % tackled by \citet{yoo2024layerwise}.
Additionally, such mechanisms may also incorporate second order optimization, intermediate objectives and dynamic learning rates to allow for a smooth learning process.

\subsubsection*{Acknowledgements}
% These acknowledgements do not count towards the page limit.
We thank Matthias De Lange, Eli Verwimp and anonymous reviewers for useful comments.
This project has been supported by funding from the European Union under the Horizon 2020 research and innovation program (ERC project KeepOnLearning, grant agreement No.\ 101021347) and under Horizon Europe (Marie Skłodowska-Curie fellowship, grant agreement No.\ 101067759).

\bibliographystyle{apalike}
\bibliography{references}

\appendix
\renewcommand\thefigure{\thesection.\arabic{figure}}  
\renewcommand\thetable{\thesection.\arabic{table}}
\section*{APPENDIX}
The Appendix contains additional information for the content presented in the main body. Appendices~\ref{apx:min_acc} and~\ref{apx:method_details} expand upon the average minimum accuracy metric utilized in the main text to quantify the stability gap, and provide the details on the BiC and DER continual learning mechanisms. Appendix~\ref{apx:ergem_implmentation} details the nuances of our ER\,+\,GEM combination approach implementation. Appendix~\ref{sec:gem_memory_strength} contains additional experiments beyond our original experimental protocol that analyze the influence of GEM's hyperparameter $\gamma$.
Finally, in Appendix~\ref{apx:additional_results}, we disclose the results for all experiments including those not already featured in the main text. 
%In summary the overall structure of the Appendix is:
% \begin{itemize}
%     \item[\textbf{A}] Average Minimum Accuracy
%     \item[\textbf{B}] Method Details BiC and DER
%     \item[\textbf{C}] Implementation of ER + GEM
%     \item[\textbf{D}] Additional Results
% \end{itemize}

Documented code to reproduce all experiments is publicly available at \url{https://github.com/TimmHess/TwoComplementaryPerspectivesCL}.

%%%%%%%%%%%%%%%%%%%%%%%%%%%%%%%%%%%%%%%%%%%%

\section{ACCURACY METRICS}
\label{apx:min_acc}
\setcounter{figure}{0}
\setcounter{table}{0}

% Details on Accuracy
\textbf{Classification accuracy} is the base metric we use throughout this work. Given a data set~$D$ and model~$f$, the classification accuracy $\textbf{A}(D, f)$ is the percentage of samples in~$D$ that is correctly classified by~$f$.

% Details on final average accuracy
\textbf{Final average accuracy} is the classification accuracy averaged over all tasks at the end of training. Formally:
%is the classification accuracy averaged over all evaluation sets $\hat{T_t}, \forall t \in T$ using the final model $f_w$, after concluding learning all $T$ tasks:
\begin{equation}
    \textbf{avg-ACC} = \frac{1}{T} \sum_{t=1}^T \mathbf{A}(\hat{D_t}, f_{w_{\text{final}}}),
\end{equation}
where $\hat{D_t}, \forall t \in [1,...,T]$ is the evaluation set of each task, and $f_{w_{\text{final}}}$ is the final model after concluding training on the all $T$~tasks. This is a common metric to express the quality of the continually learned model.

% Details on average minimum accuracy 
\textbf{Average minimum accuracy} denotes the average of the lowest classification accuracies observed for each task, as measured from directly after finishing training on that task until the end of all learning in the task sequence: %($n = |T_T|$). 
\begin{equation}
    \textbf{min-ACC}(T_t) = \min_{|T_t| < n \leq |T_T|} \mathbf{A}(\hat{D}_t, f_{w_n}),
\end{equation} 
\begin{equation}
    \textbf{avg-min-ACC} = \frac{1}{T-1} \sum_t^{T-1} \textbf{min-ACC}(T_t),
\end{equation}
where $f_{w_n}$ indicates the model after the $n^{\text{th}}$ training iteration. By slight abuse of notation, $n = |T_t|$ refers to the last training iteration of task $T_t$, and $n = |T_T|$ marks the final training iteration of the entire task sequence $\mathcal{T}$. This metric is a worst case measure of how well the model's classification accuracy is maintained at any point throughout continual training.
To render per-iteration calculation of $\mathbf{A}(\hat{D}_t, f_{w_n})$ computationally feasible, we resort to a reduced evaluation set of size $1000$ per task. The reduced set is sampled uniformly from each evaluation set respectively, once for every run. This is similar to \citet{delange2023continual} and has empirically shown to closely approximate using the full evaluation set.

%%%%%%%%%%%%%%%%%%%%%%%%%%%%%%%%%%%%%%%%%%%%

\section{METHOD DETAILS}
\label{apx:method_details}
\setcounter{figure}{0}
\setcounter{table}{0}

% Details on GEM
%\textbf{GEM} (\textbf{G}radient \textbf{E}pisodic \textbf{M}emory)\\

% Details on BiC
\textbf{BiC} (\textbf{Bi}as-\textbf{C}orrection) was proposed by \citet{wu2019large}, who were motivated by the observation that class-incremental continual training causes the continually trained classifier to become biased towards the most recently observed set of classes. The approach is specifically designed for class-incremental learning settings where each observed task $T_t$ introduces a set of non-overlapping classes $m$, such that the corresponding data $X_t^m = \{(x_i,y_i), \forall y_i \in [n+1,...,n+m]\}$. Here, $x_i, y_i$ denote the example and label pair, and $n$ is the number of already observed classes. To correct the bias, a (small) set of validation data for each task is stored. These data are taken from the training set and excluded from training the model. They are only used for correcting the classifier's bias in a separate training stage. The bias-correction itself is realized by a linear layer that consists of two parameters, $\alpha$ and $\beta$, and is called the bias-correction layer. The logits produced by the model for previously observed classes are kept unaltered, but the bias in the logits produced for the $m$ newly observed classes ($n+1,...,n+m$) is corrected by the bias-correction layer:
\begin{equation}
    q_k = 
    \begin{cases} 
    o_k &  1 \leq k \leq n \\
    \alpha o_k + \beta &  n+1 \leq k \leq n+m,
  \end{cases}
\end{equation}
where $o_k$ denotes the output logits for the $k^{\text{th}}$ class.
The bias-correction parameters ($\alpha, \beta$) are shared for all new classes and optimized via a cross-entropy classification loss:
\begin{equation} 
    L_b = - \sum_{k=1}^{n+m} \log[p_k(q_k)],
\end{equation}
with $p_k(.)$ indicating the output probability, i.e. softmax of the logits.\\

Next to the bias-correction, BiC uses data augmentation, a replay mechanism, and a distillation mechanism to continually train the model. 
The data-augmentation comprises random cropping with scales ranging $0.2$ to $1.0$ and random horizontal flipping with chance of $p=0.5$. These augmentations are also applied to buffered samples during replay.
The general replay mechanism is discussed in Section~\ref{sec:method} of the main body of this paper. Here, we simplify it to allocating an auxiliary buffer $\hat{X}$ of size $M$ that allows to interleave training with exemplars from previously observed $n$ classes. A notable addition is that this buffer holds both, the replay exemplars for training and for validation, with the latter already including samples from the current task. \citet{wu2019large} found a allocation ratio of $9:1$ for training/validation to be sufficient. The replay interleaved cross-entropy training loss is formulated as:
\begin{equation} \label{eq:replay_ce}
    L_c = \sum_{(x,y) \in \hat{X}^n\cup X^m_t} \sum_{k=1}^{n+m} - \delta_{y=k} \log(p_k(x)).
\end{equation}
The additional regularizing distillation loss is formulated as: 
\begin{equation}
    L_d = \sum_{x \in \hat{X}^n \cup X^m} \sum_{k=1}^n -\hat{\pi}_k(x) \log [\pi_k(x)],
\end{equation}
\begin{equation*}
    \hat{\pi}_k = \frac{e^{\hat{o}^n_k(x)/T}}{\sum_{j=1}^n e^{\hat{o}^n_j(x)/T}}, \ \ \  \pi_k(x) = \frac{e^{o^{n+m}_k(x)/T}}{\sum_{j=1}^n e^{o^{n+1}_j(x)/T}},
\end{equation*}
with $\hat{o}^n$ denoting the logits from the previous, old, model and $T$ the temperature scalar. Note that the previous bias-correction is applied in $\hat{o}^n$. 
Ultimately, both losses are combined to the total training loss:
\begin{equation}
    L = \lambda L_d + (1-\lambda) L_c,
\end{equation}
with balancing scalar $\lambda = \frac{n}{n+m}$, with $n$ and $m$ being the number of old and new classes respectively.
\ \\

% Details on DER
\textbf{DER} (\textbf{D}ark \textbf{E}xperience \textbf{R}eplay)
is an experience replay approach that extents the standard replay formulation~(\textit{c.f.}~Eq.~\ref{eq:replay_ce}) by a regularization term based on distillation \citep{hinton2015distilling}:  
\begin{equation}
    L_{\text{DER}} = L_c + \alpha \mathbb{E}_{(x,o_k) \sim \hat{X}} [ \text{D}_{\text{KL}}(p_k(o_k) || f(x)) ],
\end{equation}
with loss discounting hyperparameter $\alpha$. 
The auxiliary replay buffer is defined as:
$$\hat{X} = \{(x_i,o_i), 0 \leq i < M \},$$ containing $M$ pairs $(x_i, o_i)$ of previous tasks exemplars $x_i$ along with the models output logits $o_i$ (at the time of adding them to the buffer), instead of targets $y_i$.\\
Further, to avoid information loss in the softmax-function when comparing model output $f(x)$ to $o_i$, the authors chose to approximate the KL divergence (D$_{\text{KL}}$ by the Euclidean distance. With that, the final loss becomes:
\begin{equation}
    L_{\text{DER}} = L_c + \alpha \mathbb{E}_{(x,z) \sim \hat{X}} [ \ ||z-h(x)||_2^2 \ ].
\end{equation}
% Data augmentation
As for BiC, data augmentation by random cropping with scales ranging $0.2$ to $1.0$ and random horizontal flipping with chance of $p=0.5$ are applied to all forwarded data.

%%%%%%%%%%%%%%%%%%%%%%%%%%%%%%%%%%%%%%%%%%%%

\section{IMPLEMENTATION DETAILS}
\label{apx:ergem_implmentation}
\setcounter{figure}{0}
\setcounter{table}{0}

\begin{algorithm}[t]
    \caption{ER\,+\,\mbox{AGEM} (detailed)}\label{algo:eragem}
    \begin{algorithmic}[1]
        \Require{parameters $w$, loss function $\ell$, learning rate $\lambda$, 
        \phantom{for} \phantom{for} \phantom{for} \,\,\, 
        data stream $\{D_1, ..., D_T\}$}
        \vspace{0.04in}
        \State $M \gets \{\}$
        \For{$t=1,...,T$}
            \For {$(x,y) \in  D_t$} 
                \State \#1. Sample from memory buffer
                \State $(\tilde{x}_k,\tilde{y}_k) \gets \texttt{SAMPLE}(M)$
                \State
                \State \#2. Compute gradients of (approx.) joint loss
                \State $[z_{x}, z_{\tilde{x}}] \gets f_w([x, \tilde{x}])$ \label{algagem:line:forward}
                \State $g \gets \nabla_w \ell(z_x, y)$ 
                \State $g_{\text{old}} \gets \nabla_w \ell(z_{\tilde{x}}, \tilde{y})$
                \State $g_{\text{joint}} \gets \frac{1}{t}g + (1-\frac{1}{t})g_{\text{old}}$ 
                \State
                \State \#3. Compute reference gradients
                \State $g_{\text{ref}} \gets g_{\text{old}}$ 
                \State
                \State \#4. Gradient projection
                \State $\bar{g} \gets \texttt{PROJECT\_AGEM}(g_{\text{joint}}, g_{\text{ref}})$ 
                \State
                \State \#5. Update model parameters
                \State $w \gets \texttt{OPTIMIZER\_STEP}(w,\lambda,\bar{g})$
            \EndFor
            \State $M \gets \texttt{UPDATE\_BUFFER}(M, D_t)$ 
        \EndFor
    \end{algorithmic}
\end{algorithm}

\begin{algorithm}[t]
    \caption{ER\,+\,\mbox{GEM} (detailed)}\label{algo:ergem}
    \begin{algorithmic}[1]
        \Require{parameters $w$, loss function $\ell$, learning rate $\lambda$, 
        \phantom{for} \phantom{for} \,\,\, 
        hyperparameter $\gamma$, data stream $\{D_1, ..., D_T\}$}
        \vspace{0.04in}
        \State $M_t \gets \{\}$, $\forall$ $t = 1, ..., T$
        \For{$t=1,...,T$}
            \For {$(x,y) \in  D_t$}
                % sample replay data
                \State \#1. Sample from memory buffer
                \State $(\tilde{x}_k,\tilde{y}_k) \gets \texttt{SAMPLE}(M_k)$ for all $k<t$ \label{alg2:line:gem_sample}
                \State $\overline{M} \gets \{(\tilde{x}_k,\tilde{y}_k)\}$ for all $k<t$ \label{alg2:line:replay_sample_start}
                \State $(\tilde{x}_{k<t},\tilde{y}_{k<t}) \gets \texttt{SAMPLE}(\overline{M})$ \label{alg2:line:replay_sample_end}
                \State
                \State \#2. Compute gradients of (approx.) joint loss
                % forward [new,old] together
                \State $[z_{x}, z_{\tilde{x}_{k<t}}] \gets f_w([x, \tilde{x}_{k<t}])$ \label{alg2:line:forward}
                % compute gradient of joint loss
                \State $g \gets \nabla_w \ell(z_x, y)$ 
                \State $g_{\text{old}} \gets \nabla_w \ell(z_{\tilde{x}_{k<t}}, \tilde{y}_{k<t})$
                \State $g_{\text{joint}} \gets \frac{1}{t}g + (1-\frac{1}{t})g_{\text{old}}$ 
                \State
                \State \#3. Compute reference gradients
                % compute reference gradients
                \State \texttt{FREEZE\_BATCH\_NORM}($f_w$) \label{alg2:line:freeze}
                \State $g_{\text{ref}_k} \gets \nabla_w \ell(f_w(\tilde{x}_k), \tilde{y}_k)$ for all $k<t$
                \State \texttt{UNFREEZE\_BATCH\_NORM}($f_w$) \label{alg2:line:unfreeze} 
                \State
                \State \#4. Gradient projection
                % projection
                \State $\bar{g} \gets \texttt{PROJECT\_GEM}(g_{\text{joint}},[g_{\text{ref}_1}, ..., g_{\text{ref}_k}],\gamma)$ \label{alg:line:project_gem}
                \State
                \State \#5. Update model parameters
                \State $w \gets \texttt{OPTIMIZER\_STEP}(w,\lambda,\bar{g})$
            \EndFor
            \State $M_t \gets \texttt{FILL\_BUFFER}(D_t)$ 
        \EndFor
    \end{algorithmic}
\end{algorithm}

The practical implementation of combining experience replay (ER) and the gradient projection mechanism of GEM or \mbox{A-GEM} includes multiple aspects that benefit from additional clarification. 
In this section we discuss the calculation of the reference gradients and the handling of batch normalization. %, by our choice of implementation delineated in Algorithm~\ref{algo:ergem}.
To accompany this discussion, we provide detailed pseudocodes for our implementations of \eragem (Algorithm~\ref{algo:eragem}) and \ergem (Algorithm~\ref{algo:ergem}), complementing the higher-level pseudocode for \eragem in the main text. 

% 0. Replaying all data for GEM
\textbf{Calculation of reference gradients:}
In the GEM and \mbox{A-GEM} mechanism, reference gradients inform the gradient projection by indicating the gradient direction that decreases the loss on previously learned tasks.
% AGEMs reference gradients
When combining ER with \mbox{A-GEM}, we take the reference gradients to be the same as the gradients that are used in optimizing the approximate joint loss  (i.e.,~$g_{\text{ref}}=g_{\text{old}}$). Computing the reference gradients on a separately sampled mini-batch from the memory buffer may have beneficial effects for training, but would come at an increased computational cost of effectively doubling the replay mini-batch size.
When combining ER with GEM, obtaining the reference gradients is more complex because a separate reference gradient is required for each previously learned tasks. 
In the original formulation of GEM by \cite{lopez2017gradient}, the reference gradients are calculated with respect to the entire memory buffer. This quickly becomes computationally very costly if large amounts of data are stored. %, and especially if all data are stored (i.e. incremental joint training). 
As a mitigation, rather than using each task's entire buffer to compute the reference gradient, we sample one mini-batch per previously observed task. In particular, when training on task~$t$, we sample $k=t-1$~mini-batches, one from each task-specific memory buffer~$M_k$, see Algorithm~\ref{algo:ergem} line~\ref{alg2:line:gem_sample}.
All sampled mini-batches are of the same size as the mini-batch currently observed by the model. %This approximation to the individual tasks' reference gradients aligns with our formulation of `full replay' in Section~\ref{sec:method} of the main text. 
In order to closely approximate the `same-mini-batch' relation between the replay gradient and the reference gradients, % -- similar to \mbox{A-GEM --} 
we then obtain the replay mini-batch by uniformly sampling from the $k$ mini-batches of the reference gradients, see Algorithm~\ref{algo:ergem} \mbox{lines~\ref{alg2:line:replay_sample_start}-\ref{alg2:line:replay_sample_end}}.
\textbf{Batch norm:} When implementing \eragem or ER\,+\,GEM, another aspect requiring careful consideration is the use of batch normalization \citep{ioffe2015batch}, which is included in the reduced \mbox{ResNet-18} architecture that we use for all benchmarks except Rotated MNIST. %such as for calculating of the reference gradients.
When training with batch norm, the normalization statistics are computed relative to each individual mini-batch that is forwarded through the model. This means that when the current data and the replay data are forwarded through the model in different mini-batches, they use different normalization statistics, which might induce instability in the training.
%Having separate forward passes for calculations on the current and replay data induces a degree of instability to the model updates. 
%Because each batch contains data exclusive to only one task's distribution, each batch receives different normalization due to its bias.
%This is because the individually forwarded batches are biased, as each contains data exclusive to only one task's distribution, and thus receive different normalization.
%By default, batch-normalization layers accumulate an internal approximation to the global mean and variance of the observed data as a running statistic that is updated in every forward step of the model. Having separate forward passes for calculations on current and replayed data induces a degree of instability to the batch-norm. This is because the individually forwarded batches are biased, as each contains data exclusive to only one task's distribution. 
For standard replay, as well as for ER\,+\,\mbox{A-GEM}, we can mitigate this potential instability by forwarding the current and replayed data together, see Algorithm~\ref{algo:eragem} line~\ref{algagem:line:forward}. However, when using replay in combination with GEM, it becomes more complex, because more data from the memory buffer is forwarded through the model than required for approximating the joint loss. Forwarding all data together seems undesirable as it would bias the normalization statistics too much toward the data from previous tasks (and doing so might also be impractical as the mini-batch size might become too large for forwarding all data together). Instead, common implementations of GEM typically forward the data of each past task separately, which means that 
%the fact that the batch-norm layers are updating their internal variables in the forward passes 
%and in turn are implicitly introducing extra information to the batch-norm's running statistics, however biased. To not affect the batch-norm we calculate the reference gradients with frozen batch-norm after the replayed and current batch have been forwarded, see Algorithm~\ref{algo:ergem} lines~\ref{alg2:line:freeze}~-~\ref{alg2:line:unfreeze}.\\
each reference gradient is computed with a custom, task-specific normalization.
Such a different normalization for each reference gradient might induce instability in the training. To try to mitigate this, when computing the reference gradients for GEM, we freeze the batch-norm layers, see Algorithm~\ref{algo:ergem} line~\ref{alg2:line:freeze}-\ref{alg2:line:unfreeze}.

\section{ADDITIONAL STUDY OF THE OPTIMIZATION ROUTINE OF GEM}
\label{sec:gem_memory_strength}
\setcounter{figure}{0}
\setcounter{table}{0}

Here, we take a detailed look at the optimization routine of GEM; in particular, at its gradient projection step (i.e.,~line~\ref{alg:line:project_gem} in Algorithm~\ref{algo:ergem}). While performing the experiments for this paper, we realized that the optimization routine of GEM is not unambiguously defined and that it has an influential hyperparameter. The effect of this hyperparameter, which is not present in the optimization routine of A-GEM, can explain for a large part the difference in performances that we observed between using the optimization routine of GEM versus that of A-GEM (see section~\ref{sec:results} in the main text).

The motivation behind the optimization mechanism of GEM is to allow only such gradient updates that do not increase the loss on any previous task. Mathematically, \citet{lopez2017gradient} formulated GEM's optimization mechanism as follows. When training on task~$t$, the  gradient~$\bar{g}$ based upon which the optimization step is taken is given by the solution to:
\begin{align}
    \text{minimize}_{\bar{g}}& \ \ \frac{1}{2}||g-\bar{g}||_{2}^{2}\label{eq:qp1}\\
    \text{subject to}& \ \ \langle \bar{g},g_k \rangle \geq 0, \forall k < t,\label{eq:qp2}
\end{align}
where $g$ is the gradient of the loss being optimized and $g_k$ is the reference gradient computed on stored data for the $k^{\text{th}}$~task. 
Equations~(\ref{eq:qp1}) and~(\ref{eq:qp2}) define a quadratic program~(QP) in $p$ variables, with $p$ the number of trainable parameters of the neural network. To solve this QP, GEM uses the dual problem, which is given by:
\begin{align}
    \text{minimize}_{v}& \ \ \frac{1}{2}v^{\text{T}} GG^{\text{T}}+ g^{\text{T}}G^{\text{T}}v\\
    \text{subject to}& \ \ v \geq 0, \label{eq:qp_constraint}
\end{align}
with \mbox{$G = (g_1,...,g_{t-1})$}.\footnote{In the published version of \cite{lopez2017gradient}, $G$ is erroneously defined as \mbox{$-(g_1,...,g_{t-1})$}. In September 2022, the authors corrected this in the ArXiv version.} %the reference gradients computed on data from previous tasks that is stored in auxiliary buffers. 
This dual problem is a QP in only $t-1$~variables, and can therefore be solved more efficiently.
After solving this dual problem, $\bar{g}$ can be recovered as:
\begin{align} \label{eq:recover_g}
     \bar{g} = G^{\text{T}}v^\ast + g,
\end{align}
where $v^\ast$ is the solution to the dual problem.

\begin{figure}[!b]
    \centering
    \includegraphics[width=0.49\textwidth]{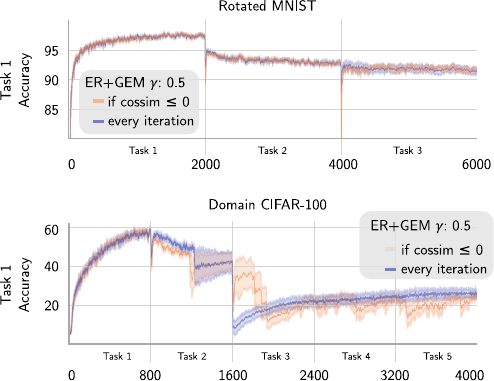}
    \caption{\textbf{When to perform the GEM projection.} Illustrated is the difference for \ergem between performing the GEM projection operation at every iteration \textit{versus} only when the cosine-similarity of the current gradient with at least one reference gradient is $\leq 0$.}
    \label{fig:gem_every_iteration}
\end{figure}

\begin{figure*}[!t]
    \centering
    \begin{tabular}{p{0.49\textwidth} p{0.49\textwidth}}
         \includegraphics[width=0.46\textwidth]{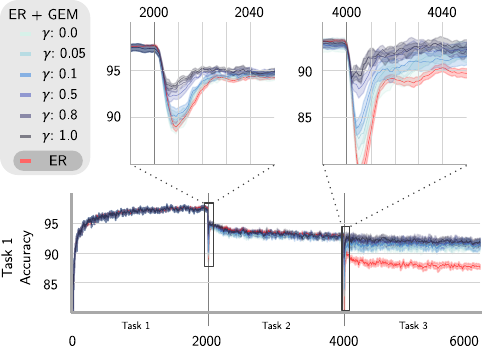} & \includegraphics[width=0.42\textwidth]{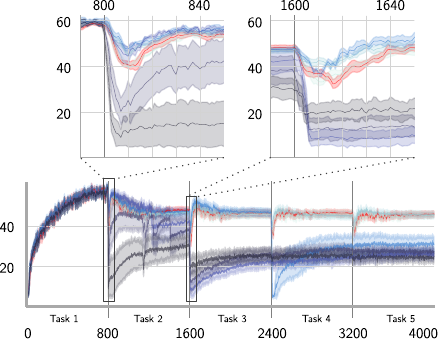}\\[1em]
        %\hspace*{1em} 
        \resizebox{0.49\textwidth}{!}{
\begin{tabular}{p{2em} |
p{3.8em}
p{3.8em} p{3.8em} p{3.8em} p{3.8em} p{3.8em} |
p{3.8em} p{0em}}
    \cmidrule{1-8}
    & \multicolumn{6}{c|}{ER + GEM} & \centering ER &\\
    \cmidrule{1-8}
    $\gamma$ & \centering $0.0$ & \centering $0.05$ & \centering $0.1$ & \centering $0.5$ & \centering $0.8$ & \centering $1.0$ & \centering $-$ &\\
    \cmidrule{1-8}
    MIN & $84.4 \pm 0.7$ & $85.6 \pm 0.5$ & $86.8 \pm 0.5$ & $89.1 \pm 0.2$ & $89.5 \pm 0.4$ & $\pmb{90.0 \pm 0.2}$ & $83.1 \pm 0.5$ \\
    AVG & $93.6 \pm 0.1$ & $93.7 \pm 0.2$ & $93.8 \pm 0.2$ & $93.9 \pm 0.1$ & $94.1 \pm 0.1$ & $\pmb{94.1 \pm 0.2}$ & $91.9 \pm 0.1$ \\
     \cmidrule{1-8}
\end{tabular}
}
&
        %\hspace*{1em} 
        \resizebox{0.49\textwidth}{!}{
\begin{tabular}{p{2em} |
p{3.8em}
p{3.8em} p{3.8em} p{3.8em} p{3.8em} p{3.8em} |
p{3.8em} p{0em}}
    \cmidrule{1-8}
    & \multicolumn{6}{c|}{ER + GEM} & \centering ER &\\
    \cmidrule{1-8}
    $\gamma$ & \centering $0.0$ & \centering $0.05$ & \centering $0.1$ & \centering $0.5$ & \centering $0.8$ & \centering $1.0$ & \centering $-$ &\\
    \cmidrule{1-8}
    MIN & $\pmb{33.0 \pm 0.6}$ & $\pmb{33.6 \pm 1.5}$ & $~~8.4 \pm 1.5$ & $~~9.0 \pm 1.0$ & $10.0 \pm 1.1$ & $10.7 \pm 1.8$ & $\pmb{33.2 \pm 1.3}$\\
    AVG & $\pmb{48.1 \pm 0.7}$ & $47.2 \pm 0.7$ & $29.0 \pm 4.9$ & $21.3 \pm 1.2$ & $22.9 \pm 0.9$ & $21.3 \pm 2.3$ & $\pmb{48.6 \pm 0.5}$ \\
     \cmidrule{1-8}
\end{tabular}
}

    \end{tabular}
    \caption{\textbf{Influence of GEM's hyperparameter $\pmb{\gamma}$ on Rotated MNIST (left) and Domain CIFAR-100 (right).} 
    The middle panels show the test accuracy on the first task while the model is incrementally trained on all tasks. The top panels show zoomed in views of the first $50$ iterations after a task switch, allowing a more detailed qualitative comparison of the stability gaps. The bottom panels display tables that quantitatively compare average minimum accuracy (MIN) and final average accuracy (AVG).}
    \label{fig:gem_memory_strength}
\end{figure*}

% \begin{algorithm}
%     \caption{PROJECT\_GEM}\label{algo:project_gem}
%     \begin{algorithmic}[1]
%         \Require{gradient $g$, reference gradients $g_{\text{ref}_1},...,g_{\text{ref}_k}$, 
%         \phantom{for} \phantom{for} \phantom{for} \,\,\, 
%         memory strength $\gamma$}
%         \State $\pmb{G} \gets [g_{\text{ref}_1},...,g_{\text{ref}_k}]$
%         \State $\pmb{P} \gets \langle \pmb{G}, \pmb{G}^T \rangle$
%         \State $\pmb{P} \gets \frac{1}{2} (\pmb{P}+\pmb{P}^T)$
%         \State $q \gets \langle \pmb{G}, g \rangle * -1$ 
%         \State $\pmb{G} \gets \pmb{I}$
%         \State $h \gets \Vec{0} + \gamma$
%         \State
%         \State $v \gets \texttt{SOLVE\_QP}(\pmb{P}, q, \pmb{G}, h)$
%     \end{algorithmic}
% \end{algorithm}

This is however not the full story. \citet{lopez2017gradient} further introduced a hyperparameter~$\gamma$, because in practice they found that \emph{`adding a small constant $\gamma \geq 0$ to $v^{\ast}$ biased the gradient projection to updates that favored beneficial backward transfer’} (p.~4). 
Based on this description, the reader might expect Equation~(\ref{eq:recover_g}) to change to $\tilde{g} = G^{\text{T}}(v^\ast+\gamma) + g$, but in the official code implementation of GEM,\footnote{\url{https://github.com/facebookresearch/GradientEpisodicMemory}} %, that is typically copied, as in the case of the Avalanche continual learning library.
$\gamma$ %, referred to as \textit{margin} or \textit{memory strength}, 
is instead added to the right-hand side of the inequality constraint of the dual problem (i.e.,~the inequality in Equation~(\ref{eq:qp_constraint}) changes to $v \geq \gamma$).

Furthermore, setting $\gamma > 0$ introduces another subtlety, because it makes that the solution $\bar{g}$ to the dual problem is always different from~$g$, even if the constraint $\langle g,g_k \rangle \geq 0$ is satisfied for each past task~$k$. Nevertheless, in the official code implementation of GEM, $\bar{g}$ is still set to $g$ if $\langle g,g_k \rangle \geq 0$ for all $k<t$. In other words, the hyperparameter~$\gamma$ is used only if $\langle g,g_k \rangle < 0$ for at least one past task~$k$. This thus introduces a discontinuity (see Figure~\ref{fig:gem_every_iteration} for an empirical evaluation of the effect of this).

The way that hyperparameter~$\gamma$ is treated in GEM's official code implementation is typically taken over by other publicly available implementations of GEM. 
%\textcolor{red}{(e.g.,~TO ADD)}.
For the pre-registered experiments reported in the main text, we followed the official code implementation of GEM as well, and we used $\gamma=0.5$, as this is the value that \citet{lopez2017gradient} used for all their main experiments. 

In this Appendix we report additional experiments that explore the impact of hyperparameter~$\gamma$. First, we note that the effect of $\gamma$ can be interpreted as enlarging the influence of the reference gradients~$g_k$ on~$\bar{g}$. The reason for this is that~$\gamma$ tends to increase~$v^\ast$, and $\bar{g}$ is related to $v^\ast$ through $\bar{g}=G^{\text{T}}v^\ast+g$.
Figure~\ref{fig:gem_memory_strength} empirically evaluates the impact of varying $\gamma$ on the performance of \ergem on the offline versions of Rotated MNIST and Domain CIFAR-100. For these experiments, the discontinuity regarding hyperparameter~$\gamma$ is removed (i.e.,~the dual problem is solved at every iteration, we do not first check whether $\langle g,g_k \rangle < 0$ for at least one $k<t$). 
For Rotated MNIST, we find that increasing $\gamma$ leads to both a reduction in the stability gap and an increase in final performance.
%Leaving $\gamma = 0$ corresponds to projecting the gradient $g$ to the constraints as defined in Eq.~\ref{eq:qp2}, the setting adopted in A-GEM.
For Domain CIFAR-100, we find that the lower~$\gamma$, the later the collapse appears, and with $\gamma\leq0.05$ we no longer observe a collapse.

%%%%%%%%%%%%%%%%%%%%%%%%%%%%%%%%%%%%%%%%%%%%

\section{ADDITIONAL RESULTS}
\label{apx:additional_results}
\setcounter{figure}{0}
\setcounter{table}{0}

\begin{figure*}[!b]
    \centering
    \includegraphics[width=0.99\textwidth]{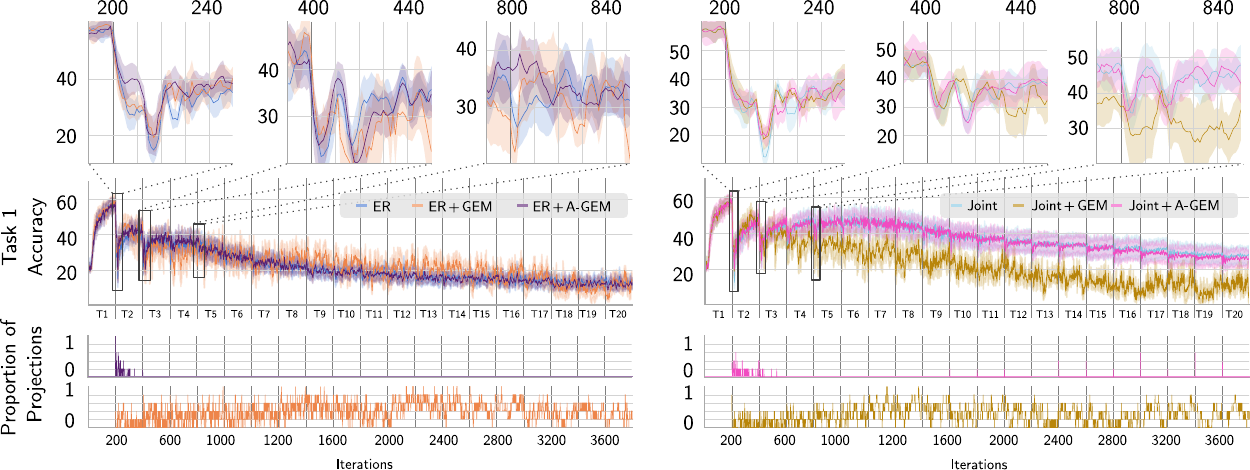}
    \caption{\textbf{Stability gaps for first task of offline Split Mini-ImageNet.} The left side shows standard ER, the right side incremental joint training (or `full replay') -- both by themselves and in combination with the optimization mechanism of GEM and A-GEM. The middle panels show the test accuracy on the first task while the model is incrementally trained for all tasks of the benchmark. The top panels show zoomed-in views of the first 50 training iterations after a task switch, allowing a more detailed qualitative comparison of the stability gap. These plots show the mean $\pm$ standard error (shaded area) over five runs with different random seeds. The bottom panel shows for every iteration the proportion of runs where the gradient was projected, with $0$ indicating that at this iteration there was no run in which a gradient was projected and~$1$~indicating that there was a gradient projection in every run.}
    \label{fig:miniimgnet_offline}
\end{figure*}

\begin{figure*}[!b]
    \centering
    \includegraphics[width=0.99\textwidth]{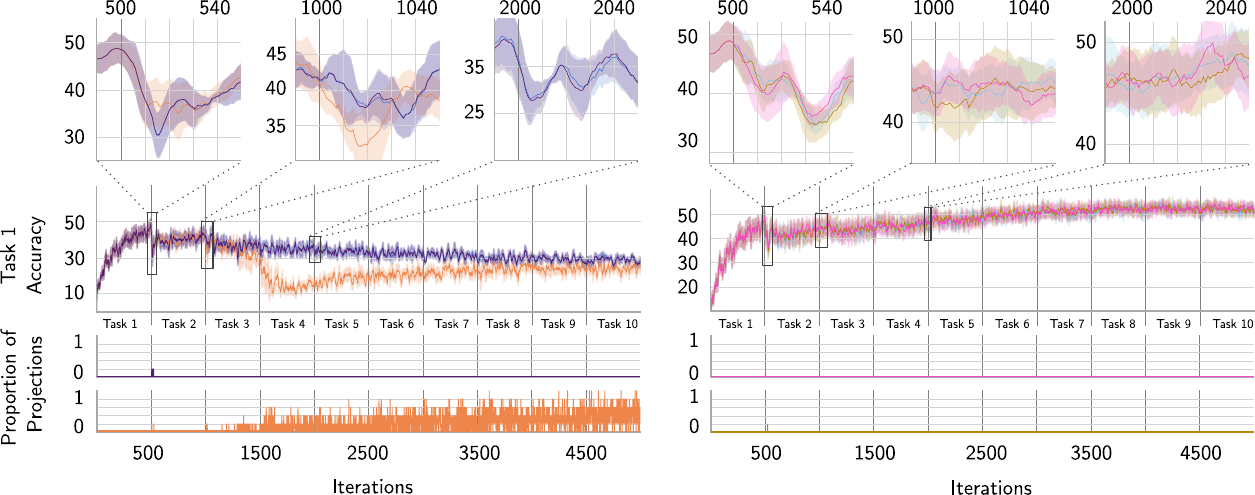}
    \caption{\textbf{Stability gaps for the first task of online Split CIFAR-100.} The left side shows standard ER, the right side incremental joint training (or `full replay') -- both by themselves and in combination with the optimization mechanism of GEM and A-GEM. The middle panels show the test accuracy on the first task while the model is incrementally trained for all tasks of the benchmark. The top panels show zoomed-in views of the first 50 training iterations after a task switch, allowing a more detailed qualitative comparison of the stability gap. These plots show the mean $\pm$ standard error (shaded area) over five runs with different random seeds. The bottom panel shows for every iteration the proportion of runs where the gradient was projected, with $0$ indicating that at this iteration there was no run in which a gradient was projected and~$1$~indicating that there was a gradient projection in every run.}
    \label{fig:cif100_online}
\end{figure*}

In this section we provide the remaining results we obtained from our experimental protocol but did not include in the main text to avoid clutter. An extensive overview of all data can be found in the Table~\ref{apx_table:rotmnist} for Rotated MNIST, Table~\ref{apx_table:domcif100} for Domain CIFAR-100, Table~\ref{apx_table:cif100} for Split CIFAR-100, and Table~\ref{apx_table:miniimg} for Mini-ImageNet. Also, we accompany the tabular view by additional plots for qualitative assessment (Figures~\ref{fig:miniimgnet_offline} and~\ref{fig:cif100_online}).

\cleardoublepage

\begin{sidewaystable*}
    \centering
    \caption{\textbf{Final average accuracy (AVG) and average minimum accuracy (MIN) for all hyperparameter settings of online and offline Rotated MNIST.} Runs marked with gray background are used for comparisons in the main body. Results reported as mean $\pm$ standard error over 5 runs with different random seeds.}
    \vspace{1em}
    \textbf{Offline}\\[0.5em]
    \resizebox{\textwidth}{!}{
\begin{tabular}{p{2em}p{2em} | p{4em}p{4em}p{5em} | p{4em}p{4em} p{5em} | p{4em}p{5em} | p{4em}p{4em}p{4em}p{0em}}
\toprule
% & Metric & Finetune & GEM & AGEM & Replay & Replay GEM & Replay AGEM \\
\vspace{-0.25em} \centering LR & Final ACC  %
& \vspace{-0.25em} \centering Finetune & \vspace{-0.25em} \centering GEM & \vspace{-0.25em} \centering AGEM%
& \vspace{-0.25em} \centering ER & \centering ER \newline \mbox{+ GEM} & \centering ER \mbox{+ AGEM}% 
& \vspace{-0.25em} \centering DER$^\ast$ & \centering DER$^\ast$ \mbox{+ AGEM}
& \centering \vspace{-0.25em} Joint & \centering Joint \mbox{+ GEM} & \centering Joint \mbox{+ AGEM} & %
\\
\midrule
% LR 0.1
\multirow{2}{*}{$0.1$} 
&  MIN 
&  $20.9 \pm 0.8$ &  $43.6 \pm 3.7$ & $33.0 \pm 1.3$ & \cc $83.1 \pm 0.5$ & \cc $84.1 \pm 0.4$ & \cc $82.5 \pm 0.7$  &  $42.3 \pm 3.9$ &  $30.3 \pm 8.1$ & \cc $86.7 \pm 0.8$ & \cc $87.5 \pm 0.9$ & \cc $86.6 \pm 0.6$ \\
&  AVG 
& $52.8 \pm 0.6$ &  $91.8 \pm 0.2$ &  $65.2 \pm 0.8$ & \cc $91.9 \pm 0.1$ & \cc $93.7 \pm 0.1$ & \cc $91.8 \pm 0.2$ &  $54.8 \pm 4.5$ &  $44.8 \pm 9.9$ & \cc $97.5 \pm 0.0$ & \cc $97.8 \pm 0.0$ &  \cc $97.5 \pm 0.0$ \\
\midrule
\multirow{2}{*}{$0.01$} 
& MIN 
& $30.2 \pm 0.4$ & $55.9 \pm 3.0$ & $39.5 \pm 1.2$ & $82.8 \pm 0.6$ & $84.4 \pm 0.2$ & $82.8 \pm 0.7$ & \cc $82.6 \pm 0.5$ & \cc $83.0 \pm 0.4$ & $86.2 \pm 0.5$ & $87.5 \pm 0.5$ &  $86.9 \pm 0.5$ \\
& AVG 
& $56.2 \pm 0.2$ & $92.4 \pm 0.1$ & $71.1 \pm 0.5$ & $91.7 \pm 0.1$ & $93.4 \pm 0.1$ & $91.8 \pm 0.1$ & \cc $87.3 \pm 0.2$ & \cc $87.1 \pm 0.2$ &  $96.6 \pm 0.0$ & $97.0 \pm 0.0$ & $96.5 \pm 0.0$ \\
\midrule
\multirow{2}{*}{$0.001$} 
& MIN 
&  $31.2 \pm 0.5$ &  $84.6 \pm 0.4$ &  $47.7 \pm 0.7$ &  $84.9 \pm 0.3$ &  $86.5 \pm 0.4$ &  $84.9 \pm 0.3$ &  $60.2 \pm 1.1$ &  $60.2 \pm 1.1$ &  $88.0 \pm 0.3$ &  $88.1 \pm 0.3$ &  $88.0 \pm 0.3$ \\
& AVG 
&  $53.2 \pm 0.3$ &  $90.8 \pm 0.1$ &  $65.4 \pm 0.4$ &  $87.6 \pm 0.2$ &  $89.4 \pm 0.1$ &  $87.6 \pm 0.2$ &  $68.6 \pm 0.4$ &  $68.6 \pm 0.4$ &  $92.0 \pm 0.1$ &  $92.1 \pm 0.0$ &  $92.0 \pm 0.0$ \\
\bottomrule
\end{tabular}
}
\\
    \vspace{2em}
    \textbf{Online}\\[0.5em]
    \resizebox{\textwidth}{!}{
\begin{tabular}{p{2em}p{2em}p{2em} | p{4em}p{4em}p{5em} | p{4em}p{4em}p{5em} | p{4em}p{5em} | p{4em}p{4em}p{4em}p{0em}}

\toprule
\vspace{-0.25em} \centering BS & \vspace{-0.25em} \centering LR & Final ACC  %
& \vspace{-0.25em} \centering Finetune & \vspace{-0.25em} \centering GEM & \vspace{-0.25em} \centering AGEM%
& \vspace{-0.25em} \centering ER & \centering ER \mbox{+ GEM} & \centering ER \mbox{+ AGEM}% 
& \vspace{-0.25em} \centering DER$^\ast$ & \centering DER$^\ast$ \mbox{+ AGEM}
& \centering \vspace{-0.25em} Joint & \centering Joint \mbox{+ GEM} & \centering Joint \mbox{+ AGEM} & %
\\
\midrule

%%%%%%%%%%%%%%%%%%%%%%%%%% 
% BS 10
%%%%%%%%%%%%%%%%%%%%%%%%%% 

% LR 0.1
\multirow{6}{*}{$10$} & \multirow{2}{*}{$0.1$} 
& \text{MIN} 
&  $~~9.8 \pm 0.0$ &   $~~9.8 \pm 0.0$ &   $~~9.8 \pm 0.0$ &   $~~9.8 \pm 0.0$ &   $~~9.8 \pm 0.0$ &   $~~9.8 \pm 0.0$ &   $~~7.6 \pm 0.4$ &   $~~8.0 \pm 0.2$  &   $~~9.8 \pm 0.0$ &   $~~9.8 \pm 0.0$ &  $~~9.8 \pm 0.0$ \\
& & \text{AVG} 
&  $10.1 \pm 0.4$ &  $10.1 \pm 0.4$ &  $10.1 \pm 0.4$ &  $10.1 \pm 0.4$ &  $10.1 \pm 0.4$ &  $10.1 \pm 0.4$ &  $10.3 \pm 0.3$ &  $10.3 \pm 0.3$ &  $10.1 \pm 0.4$ &  $10.1 \pm 0.4$ &   $10.1 \pm 0.4$ \\
\cmidrule{2-14}
% LR 0.01
& \multirow{2}{*}{$0.01$} 
& \text{MIN} 
&  $23.7 \pm 0.7$ &  $79.5 \pm 2.6$ &  $36.3 \pm 1.2$ & \cc $86.8 \pm 0.3$ & \cc $89.1 \pm 0.4$ & \cc $87.1 \pm 0.4$ &  $43.8 \pm 4.5$ &  $46.7 \pm 3.5$ & \cc $92.3 \pm 0.4$ & \cc $92.8 \pm 0.3$ &  \cc $92.4 \pm 0.1$ \\
& & \text{AVG} 
&  $53.8 \pm 0.6$ &  $92.9 \pm 0.1$ &  $64.6 \pm 1.1$ & \cc $92.7 \pm 0.1$ & \cc $94.2 \pm 0.1$ & \cc $92.8 \pm 0.2$ &  $57.9 \pm 6.1$ &  $60.9 \pm 2.7$ & \cc $96.8 \pm 0.0$ & \cc $96.8 \pm 0.1$ &  \cc $96.8 \pm 0.1$ \\
\cmidrule{2-14}
% LR 0.001
& \multirow{2}{*}{$0.001$} & MIN 
&  $31.4 \pm 0.3$ &  $82.5 \pm 0.5$ &  $47.2 \pm 0.7$ &  $86.3 \pm 0.3$ &  $88.0 \pm 0.2$ &  $86.3 \pm 0.3$ &  $69.8 \pm 0.7$ &  $69.1 \pm 0.9$ &  $90.9 \pm 0.4$ &  $91.5 \pm 0.3$ &   $91.0 \pm 0.3$ \\
& & AVG 
&  $55.8 \pm 0.3$ &  $92.4 \pm 0.1$ &  $69.3 \pm 0.2$ &  $90.8 \pm 0.3$ &  $92.3 \pm 0.1$ &  $90.8 \pm 0.3$ &  $78.3 \pm 0.3$ &  $78.3 \pm 0.3$ &  $95.1 \pm 0.1$ &  $95.4 \pm 0.0$ &   $95.1 \pm 0.1$ \\

\midrule
%%%%%%%%%%%%%%%%%%%%%%%%%% 
% BS 64
%%%%%%%%%%%%%%%%%%%%%%%%%% 
% LR 0.1
\multirow{6}{*}{$64$} & \multirow{2}{*}{$0.1$} & \text{MIN  } 
&  $16.1 \pm 1.2$ &  $54.3 \pm 2.6$ &  $23.4 \pm 1.1$ &  $84.2 \pm 0.7$ &  $84.2 \pm 1.0$ &  $84.3 \pm 0.6$ &  $7.3 \pm 0.3$ &  $7.9 \pm 0.3$ &  $87.5 \pm 0.7$ &  $88.4 \pm 0.8$ &   $86.1 \pm 0.5$ \\
& & \text{AVG  } 
&  $48.0 \pm 0.6$ &  $86.9 \pm 0.8$ &  $56.3 \pm 0.7$ &  $91.2 \pm 0.2$ &  $93.0 \pm 0.3$ &  $91.0 \pm 0.1$ &  $9.9 \pm 0.1$ &  $9.9 \pm 0.1$ &  $96.2 \pm 0.1$ &  $96.2 \pm 0.1$ &   $96.1 \pm 0.0$ \\
\cmidrule{2-14}
% LR 0.01
& \multirow{2}{*}{$0.01$} & \text{MIN  } 
&  $28.9 \pm 0.3$ &  $70.0 \pm 2.0$ &  $40.6 \pm 0.8$ &  $83.9 \pm 0.4$ &  $84.6 \pm 0.5$ &  $83.9 \pm 0.3$ &  $72.5 \pm 0.4$ &  $72.1 \pm 0.5$ &  $87.6 \pm 0.6$ &  $88.8 \pm 0.6$ &   $87.3 \pm 0.5$ \\
& & \text{AVG  } 
&  $54.3 \pm 0.3$ &  $91.9 \pm 0.2$ &  $66.5 \pm 0.7$ &  $91.2 \pm 0.2$ &  $92.7 \pm 0.1$ &  $91.3 \pm 0.1$ &  $80.5 \pm 0.3$ &  $80.4 \pm 0.3$ &  $95.6 \pm 0.1$ &  $95.8 \pm 0.1$ &   $95.5 \pm 0.1$ \\
\cmidrule{2-14}
% LR 0.001
& \multirow{2}{*}{$0.001$} & MIN   
&  $30.0 \pm 0.3$ &  $79.0 \pm 0.3$ &  $39.6 \pm 0.6$ &  $81.6 \pm 0.3$ &  $82.4 \pm 0.3$ &  $81.6 \pm 0.3$ &  $42.6 \pm 0.6$ &  $42.8 \pm 0.6$ &  $83.9 \pm 0.3$ &  $83.9 \pm 0.4$ &   $84.1 \pm 0.3$ \\
& & AVG   
&  $51.4 \pm 0.2$ &  $87.9 \pm 0.2$ &  $58.6 \pm 0.5$ &  $83.7 \pm 0.3$ &  $84.9 \pm 0.2$ &  $83.8 \pm 0.3$ &  $52.1 \pm 0.4$ &  $52.2 \pm 0.4$ &  $86.5 \pm 0.1$ &  $86.6 \pm 0.1$ &   $86.5 \pm 0.1$ \\
\midrule

%%%%%%%%%%%%%%%%%%%%%%%%%% 
% BS 128
%%%%%%%%%%%%%%%%%%%%%%%%%% 
% LR 0.1
\multirow{6}{*}{$128$} & \multirow{2}{*}{$0.1$} & \text{MIN  } 
&  $20.4 \pm 0.7$ &  $39.0 \pm 3.9$ &  $27.4 \pm 0.8$ & $83.2 \pm 0.6$ & $83.6 \pm 0.6$ & $83.5 \pm 0.6$ &  $59.8 \pm 1.2$ &  $60.2 \pm 0.6$ & $87.1 \pm 0.5$ & $88.8 \pm 0.4$ & $87.2 \pm 0.8$ \\
& & \text{AVG  } 
&  $51.6 \pm 0.3$ &  $91.4 \pm 0.1$ &  $57.9 \pm 0.5$ &  $91.6 \pm 0.2$ & $93.4 \pm 0.1$ & $91.5 \pm 0.1$ &  $71.3 \pm 0.4$ &  $72.2 \pm 0.5$ & $96.5 \pm 0.0$ & $96.8 \pm 0.1$ &  $96.5 \pm 0.1$ \\
\cmidrule{2-14}
% LR 0.01
& \multirow{2}{*}{$0.01$} & \text{MIN  } 
&  $28.7 \pm 0.5$ &  $61.8 \pm 2.3$ &  $38.2 \pm 0.9$ &  $83.4 \pm 0.6$ &  $83.5 \pm 0.5$ &  $83.3 \pm 0.6$ &  $66.7 \pm 0.6$ &  $66.7 \pm 0.6$ &  $86.9 \pm 0.4$ &  $87.2 \pm 0.5$ &   $87.0 \pm 0.4$ \\
& & \text{AVG  } 
&  $53.7 \pm 0.4$ &  $90.7 \pm 0.3$ &  $62.8 \pm 0.4$ &  $90.0 \pm 0.1$ &  $91.6 \pm 0.1$ &  $90.0 \pm 0.1$ &  $76.1 \pm 0.3$ &  $75.9 \pm 0.2$ &  $94.4 \pm 0.1$ &  $94.7 \pm 0.0$ &   $94.4 \pm 0.0$ \\
\cmidrule{2-14}
% LR 0.001
& \multirow{2}{*}{$0.001$} & MIN   
&  $33.7 \pm 0.4$ &  $76.1 \pm 0.4$ &  $41.3 \pm 0.6$ &  $77.6 \pm 0.2$ &  $78.1 \pm 0.2$ &  $77.6 \pm 0.2$ &  $29.6 \pm 0.7$ &  $30.0 \pm 0.7$ &  $79.1 \pm 0.2$ &  $79.2 \pm 0.2$ &   $79.1 \pm 0.2$ \\
& & AVG   
&  $52.8 \pm 0.2$ &  $83.5 \pm 0.2$ &  $58.0 \pm 0.2$ &  $77.1 \pm 0.1$ &  $77.9 \pm 0.1$ &  $77.1 \pm 0.1$ &  $37.9 \pm 0.4$ &  $38.1 \pm 0.4$ &  $78.8 \pm 0.1$ &  $78.8 \pm 0.1$ &   $78.7 \pm 0.1$ \\
\bottomrule

\end{tabular}
}
\\
    \label{apx_table:rotmnist}
\end{sidewaystable*}

\begin{sidewaystable*}
    \centering
    \caption{\textbf{Final average accuracy (AVG) and average minimum accuracy (MIN) for all hyperparameter settings of online and offline Domain CIFAR-100}. Runs marked with gray background are used for comparisons in the main body. Results reported as mean $\pm$ standard error over 5 runs with different random seeds.}
    \vspace{1em}
    \textbf{Offline}\\[0.5em]
    \resizebox{\textwidth}{!}{
\begin{tabular}{p{2em}p{2em} | p{4em}p{4em}p{5em} | p{4em}p{4em}p{5em} | p{4em}p{5em} | p{4em}p{4em}p{4em}p{0em}}
\toprule
% & Metric & Finetune & GEM & AGEM & Replay & Replay GEM & Replay AGEM \\
\vspace{-0.25em} \centering LR & Final ACC  %
& \vspace{-0.25em} \centering Finetune & \vspace{-0.25em} \centering GEM & \vspace{-0.25em} \centering AGEM%
& \vspace{-0.25em} \centering ER & \centering ER \mbox{+ GEM} & \centering ER \mbox{+ AGEM}% 
& \vspace{-0.25em} \centering DER$^\ast$ & \centering DER$^\ast$ \mbox{+ AGEM}
& \centering \vspace{-0.25em} Joint & \centering Joint \mbox{+ GEM} & \centering Joint \mbox{+ AGEM} & %
\\
\midrule
% LR 0.1
%\multirow{2}{*}{lr $0.1$} 
& MIN 
&  $12.2 \pm 1.1$ &   $3.5 \pm 0.3$ &  $12.2 \pm 0.5$ & \cc $33.2 \pm 1.3$ &  \cc $7.2 \pm 2.6$ & \cc $34.0 \pm 0.9$ &  $46.0 \pm 1.0$ &  $46.3 \pm 1.0$ & \cc $42.9 \pm 0.7$ & \cc $42.2 \pm 1.1$ & \cc $42.7 \pm 0.8$ \\
\multirow{-2}{*}{$0.1$}
& AVG 
&  $35.9 \pm 0.7$ &  $17.6 \pm 3.1$ &  $36.1 \pm 0.9$ & \cc $48.6 \pm 0.5$ & \cc $23.9 \pm 1.8$ & \cc $48.6 \pm 0.5$ &  $59.6 \pm 1.0$ &  $60.3 \pm 0.6$ & \cc $52.4 \pm 0.8$ & \cc $52.6 \pm 0.6$ & \cc $52.0 \pm 0.7$ \\
\midrule
\multirow{2}{*}{$0.01$} 
& MIN 
&  $11.6 \pm 0.8$ &   $5.1 \pm 0.5$ &  $12.9 \pm 1.2$ &  $32.6 \pm 0.6$ &   $3.8 \pm 0.4$ &  $30.8 \pm 0.6$ & \cc $44.4 \pm 0.7$ & \cc $44.8 \pm 1.0$ &  $41.0 \pm 0.8$ &  $19.8 \pm 6.9$ &  $40.2 \pm 0.6$ \\
& AVG 
&  $36.9 \pm 0.6$ &  $18.0 \pm 1.2$ &  $37.2 \pm 0.5$ &  $46.2 \pm 0.3$ &  $16.4 \pm 1.5$ &  $46.2 \pm 0.2$ & \cc $59.7 \pm 0.5$ & \cc $60.7 \pm 0.4$ &  $50.2 \pm 0.2$ &  $42.2 \pm 6.9$ &  $50.1 \pm 0.3$ \\
\midrule
\multirow{2}{*}{$0.001$} 
& MIN 
&  $16.9 \pm 0.6$ &   $4.4 \pm 0.5$ &  $16.1 \pm 0.4$ &  $29.0 \pm 0.4$ &   $3.5 \pm 0.4$ &  $29.4 \pm 0.4$ &  $34.4 \pm 0.4$ &  $34.4 \pm 0.3$ &  $36.8 \pm 0.2$ &  $36.5 \pm 0.4$ &  $36.6 \pm 0.4$ \\
& AVG 
&  $32.2 \pm 0.4$ &  $20.4 \pm 1.2$ &  $32.5 \pm 0.3$ &  $35.0 \pm 0.2$ &  $13.7 \pm 2.5$ &  $35.6 \pm 0.3$ &  $46.2 \pm 0.3$ &  $46.0 \pm 0.3$ &  $39.6 \pm 0.1$ &  $39.4 \pm 0.2$ &  $39.3 \pm 0.4$ \\
\bottomrule
\end{tabular}
}
\\
    \vspace{2em}
    \textbf{Online}\\[0.5em]
    \resizebox{\textwidth}{!}{
\begin{tabular}{p{2em}p{2em}p{2em} | p{4em}p{4em}p{4em} | p{4em}p{4em}p{4em} | p{4em}p{4em} | p{4em}p{4em}p{4em}p{0em}}

\toprule
\vspace{-0.25em} \centering BS & \vspace{-0.25em} \centering LR & Final ACC  %
& \vspace{-0.25em} \centering Finetune & \vspace{-0.25em} \centering GEM & \vspace{-0.25em} \centering AGEM%
& \vspace{-0.25em} \centering ER & \centering ER \mbox{+ GEM} & \centering ER \mbox{+ AGEM}% 
& \vspace{-0.25em} \centering DER$^\ast$ & \centering DER$^\ast$ \mbox{+ AGEM}
& \centering \vspace{-0.25em} Joint & \centering Joint \mbox{+ GEM} & \centering Joint \mbox{+ AGEM} & %
\\
\midrule

%%%%%%%%%%%%%%%%%%%%%%%%%% 
% BS 10
%%%%%%%%%%%%%%%%%%%%%%%%%% 
% LR 0.1
\multirow{6}{*}{$10$} & \multirow{2}{*}{$0.1$} 
& \text{MIN} 
&  $11.9 \pm 0.6$ &   $~~6.6 \pm 1.3$ &  $12.1 \pm 0.4$ &  $25.6 \pm 0.9$ &   $~~5.6 \pm 1.6$ &  $25.6 \pm 0.9$ &  $15.4 \pm 0.7$ &  $14.8 \pm 0.6$ &  $29.1 \pm 0.7$ &  $30.7 \pm 1.0$ &   $30.6 \pm 1.0$ \\
& & \text{AVG} 
&  $23.8 \pm 0.8$ &  $20.9 \pm 1.8$ &  $25.1 \pm 0.6$ &  $35.1 \pm 0.8$ &  $20.7 \pm 4.7$ &  $35.1 \pm 0.8$ &  $26.0 \pm 0.8$ &  $25.5 \pm 0.8$ &  $43.2 \pm 0.5$ &  $45.3 \pm 0.9$ &   $45.0 \pm 1.0$ \\
\cmidrule{2-14}
% LR 0.01
& \multirow{2}{*}{$0.01$} 
& \text{MIN} 
&  $14.6 \pm 0.4$ &  $14.0 \pm 0.6$ &  $15.1 \pm 0.5$ & \cc $29.2 \pm 0.7$ &  \cc $~~4.3 \pm 0.1$ & \cc $29.2 \pm 0.7$ &  $20.3 \pm 0.3$ &  $20.1 \pm 0.4$ & \cc $35.5 \pm 1.1$ & \cc $35.6 \pm 1.3$ &  \cc $35.1 \pm 0.9$ \\
& & \text{AVG} 
&  $27.6 \pm 0.6$ &  $21.1 \pm 0.8$ &  $28.0 \pm 0.7$ & \cc $38.3 \pm 0.8$ & \cc $19.8 \pm 1.4$ & \cc $38.3 \pm 0.8$ &  $30.7 \pm 0.3$ &  $30.6 \pm 0.7$ & \cc $49.8 \pm 1.0$ & \cc $49.4 \pm 1.5$ & \cc  $49.7 \pm 1.2$ \\
\cmidrule{2-14}
% LR 0.001
& \multirow{2}{*}{$0.001$} 
& MIN 
&  $16.3 \pm 0.6$ &  $15.3 \pm 0.8$ &  $17.1 \pm 0.6$ &  $30.7 \pm 0.7$ &   $~~4.0 \pm 0.1$ &  $30.7 \pm 0.7$ &  $23.7 \pm 0.2$ &  $23.8 \pm 0.3$ &  $38.1 \pm 0.8$ &  $37.9 \pm 0.8$ &   $37.9 \pm 0.7$ \\
& & AVG 
&  $28.9 \pm 0.6$ &  $23.8 \pm 1.2$ &  $29.9 \pm 0.4$ &  $38.0 \pm 0.5$ &  $26.7 \pm 3.3$ &  $38.0 \pm 0.5$ &  $34.0 \pm 0.5$ &  $33.5 \pm 0.5$ &  $48.3 \pm 0.4$ &  $48.1 \pm 0.6$ &   $49.0 \pm 0.4$ \\

\midrule
%%%%%%%%%%%%%%%%%%%%%%%%%% 
% BS 64
%%%%%%%%%%%%%%%%%%%%%%%%%% 
% LR 0.1
\multirow{6}{*}{$64$} & \multirow{2}{*}{$0.1$} 
& \text{MIN  } 
&  $13.3 \pm 0.4$ &   $~~3.9 \pm 0.2$ &  $13.7 \pm 0.5$ &  $25.7 \pm 0.6$ &  $21.2 \pm 4.3$ &  $25.7 \pm 0.6$ &  $17.3 \pm 0.5$ &  $17.3 \pm 0.5$ &  $31.0 \pm 0.8$ &  $31.6 \pm 1.1$ &   $30.9 \pm 1.2$ \\
& & \text{AVG  } 
&  $24.2 \pm 0.4$ &  $12.1 \pm 2.0$ &  $24.9 \pm 0.4$ &  $34.7 \pm 1.2$ &  $29.3 \pm 4.5$ &  $34.7 \pm 1.2$ &  $26.3 \pm 0.7$ &  $26.4 \pm 0.5$ &  $45.4 \pm 1.1$ &  $44.2 \pm 1.3$ &   $44.9 \pm 1.2$ \\
\cmidrule{2-14}
% LR 0.01
& %\multirow{2}{*}{$0.01$} 
& \text{MIN} 
& $14.6 \pm 0.3$ &  $~~5.5 \pm 0.4$ & $15.4 \pm 0.5$ & $28.0 \pm 0.3$ & $27.7 \pm 1.0$ & $28.0 \pm 0.3$ & $21.5 \pm 0.5$ & $21.7 \pm 0.4$ & $37.1 \pm 1.0$ & $37.0 \pm 0.8$ & $37.0 \pm 0.9$ \\
& \multirow{-2}{*}{$0.01$} & \text{AVG} 
& $27.2 \pm 0.3$ & $19.5 \pm 1.9$ & $28.2 \pm 0.7$ & $35.9 \pm 0.4$ & $35.1 \pm 0.5$ & $35.9 \pm 0.4$ & $31.5 \pm 0.6$ & $32.2 \pm 0.8$ & $48.9 \pm 0.2$ & $49.5 \pm 0.6$ & $48.9 \pm 0.1$ \\
\cmidrule{2-14}
% LR 0.001
& \multirow{2}{*}{$0.001$} 
& MIN   
&  $16.1 \pm 0.4$ &   $~~8.6 \pm 0.9$ &  $16.2 \pm 0.3$ &  $25.3 \pm 0.3$ &   $~~6.1 \pm 2.7$ &  $25.3 \pm 0.3$ &  $19.6 \pm 0.3$ &  $20.0 \pm 0.2$ &  $27.8 \pm 0.3$ &  $22.6 \pm 3.1$ &   $27.8 \pm 0.4$ \\
& & AVG   
&  $24.4 \pm 0.5$ &  $23.4 \pm 2.5$ &  $24.8 \pm 0.6$ &  $31.5 \pm 0.2$ &  $13.9 \pm 4.4$ &  $31.5 \pm 0.2$ &  $25.9 \pm 0.5$ &  $26.1 \pm 0.6$ &  $34.4 \pm 0.2$ &  $34.5 \pm 0.2$ &   $34.5 \pm 0.2$ \\

\midrule
%%%%%%%%%%%%%%%%%%%%%%%%%% 
% BS 128
%%%%%%%%%%%%%%%%%%%%%%%%%% 
% LR 0.1
\multirow{6}{*}{$128$} & \multirow{2}{*}{$0.1$} 
& \text{MIN  } 
&  $11.8 \pm 0.5$ &   $~~3.5 \pm 0.2$ &  $11.8 \pm 0.7$ &  $24.7 \pm 0.6$ &  $20.5 \pm 4.2$ &  $24.7 \pm 0.6$ &  $17.6 \pm 0.4$ &  $17.6 \pm 0.5$ &  $29.3 \pm 1.0$ &  $29.8 \pm 0.8$ &   $29.9 \pm 0.8$ \\
& & \text{AVG  } 
&  $23.4 \pm 1.0$ &  $12.0 \pm 0.5$ &  $23.5 \pm 0.7$ &  $34.0 \pm 0.7$ &  $28.6 \pm 4.8$ &  $34.0 \pm 0.7$ &  $25.1 \pm 0.5$ &  $25.7 \pm 0.8$ &  $42.0 \pm 0.8$ &  $43.4 \pm 1.0$ &   $44.3 \pm 1.0$ \\
\cmidrule{2-14}
% LR 0.01
& \multirow{2}{*}{$0.01$} 
& \text{MIN  } 
&  $13.8 \pm 0.6$ &   $~~4.1 \pm 0.2$ &  $13.1 \pm 0.8$ &  $27.2 \pm 0.7$ &  $26.7 \pm 0.5$ &  $27.2 \pm 0.7$ &  $20.2 \pm 0.3$ &  $20.3 \pm 0.4$ &  $34.1 \pm 0.5$ &  $34.1 \pm 0.5$ &   $34.2 \pm 0.5$ \\
& & \text{AVG  } 
&  $25.9 \pm 0.6$ &  $12.5 \pm 1.8$ &  $26.0 \pm 0.9$ &  $34.7 \pm 0.4$ &  $35.3 \pm 0.8$ &  $34.7 \pm 0.4$ &  $30.2 \pm 0.4$ &  $30.1 \pm 0.3$ &  $45.1 \pm 0.5$ &  $45.0 \pm 0.6$ &   $45.0 \pm 0.7$ \\
\cmidrule{2-14}
% LR 0.001
& \multirow{2}{*}{$0.001$} 
& MIN   
&  $15.1 \pm 0.7$ &   $~~6.6 \pm 1.0$ &  $15.5 \pm 0.5$ &  $22.5 \pm 0.5$ &   $~~9.5 \pm 3.3$ &  $22.5 \pm 0.5$ &  $18.3 \pm 0.5$ &  $18.4 \pm 0.5$ &  $24.0 \pm 0.5$ &  $22.5 \pm 0.8$ &   $23.4 \pm 0.5$ \\
& & AVG   
&  $22.3 \pm 0.4$ &  $10.3 \pm 2.0$ &  $22.7 \pm 0.5$ &  $26.9 \pm 0.2$ &  $18.1 \pm 3.7$ &  $26.9 \pm 0.2$ &  $23.7 \pm 0.2$ &  $23.7 \pm 0.3$ &  $28.1 \pm 0.2$ &  $26.6 \pm 1.5$ &   $28.2 \pm 0.2$ \\
\bottomrule

\end{tabular}
}

    \label{apx_table:domcif100}
\end{sidewaystable*}

\begin{sidewaystable*}
    \centering
    \caption{\textbf{Final average accuracy (AVG) and average minimum accuracy (MIN) for all hyperparameter settings of online and offline Split-CIFAR100.} Runs marked with gray background are used for comparisons in the main body. Results reported as mean $\pm$ standard error over 5 runs with different random seeds.}
    \vspace{1em}
    \textbf{Offline}\\[0.5em]
    \resizebox{\textwidth}{!}{
\begin{tabular}{p{2em}p{2em} |
p{4em}p{4em}p{5em} |
p{4em}p{4em}p{5em} |
p{4em}p{5em} |
p{4em}p{4em} |
p{4em}p{4em}p{4em}p{0em}}
\toprule
% & Metric & Finetune & GEM & AGEM & Replay & Replay GEM & Replay AGEM \\
\vspace{-0.25em} \centering LR & Final ACC  %
& \vspace{-0.25em} \centering Finetune & \vspace{-0.25em} \centering GEM & \vspace{-0.25em} \centering AGEM%
& \vspace{-0.25em} \centering ER & \centering ER \mbox{+ GEM} & \centering ER \mbox{+ AGEM}% 
& \vspace{-0.25em} \centering DER$^\ast$ & \centering DER$^\ast$ \mbox{+ AGEM}%
& \vspace{-0.25em} \centering BiC$^\ast$ & \centering BiC$^\ast$ \mbox{+ AGEM}%
& \centering \vspace{-0.25em} Joint & \centering Joint \mbox{+ GEM} & \centering Joint \mbox{+ AGEM} & %
\\
\midrule
% LR 0.1
%\multirow{2}{*}{lr $0.1$} 
& MIN 
&  $~~0.0 \pm 0.0$ &  $~~0.0 \pm 0.0$ &  $~~0.0 \pm 0.0$ & \cc $12.4 \pm 0.3$ & \cc $~~0.0 \pm 0.0$ & \cc $12.1 \pm 0.3$ & \cc $~~1.4 \pm 0.1$ & \cc $~~1.6 \pm 0.2$ & \cc $22.3 \pm 0.9$ & \cc $23.1 \pm 1.1$ & \cc $22.5 \pm 0.2$ & \cc $~~2.8 \pm 1.9$ & \cc $23.1 \pm 0.4$ \\
\multirow{-2}{*}{$0.1$} 
& AVG 
&  $~~6.5 \pm 0.4$ &  $~~1.5 \pm 0.2$ &  $~~6.4 \pm 0.3$ & \cc $22.8 \pm 0.4$ & \cc $~~7.1 \pm 1.9$ & \cc $22.4 \pm 0.6$ & \cc $26.0 \pm 0.9$ & \cc $25.6 \pm 1.3$ & \cc $38.3 \pm 1.0$ & \cc $38.4 \pm 0.9$ & \cc $32.2 \pm 0.5$ & \cc $16.5 \pm 6.7$ & \cc $32.5 \pm 0.4$ \\
\midrule
\multirow{2}{*}{$0.01$} 
& MIN 
&  $~~0.0 \pm 0.0$ &  $~~0.0 \pm 0.0$ &  $~~0.0 \pm 0.0$ &  $10.8 \pm 0.6$ &  $~~0.0 \pm 0.0$ &  $10.9 \pm 0.7$ &   $~~1.0 \pm 0.3$ &   $~~1.0 \pm 0.2$ &  $27.4 \pm 0.4$ &  $27.8 \pm 0.5$ &  $21.6 \pm 0.7$ &   $~~3.8 \pm 0.7$ &  $21.3 \pm 0.8$ \\
& AVG 
&  $~~6.6 \pm 0.4$ &  $~~3.4 \pm 0.5$ &  $~~6.4 \pm 0.4$ &  $18.5 \pm 0.6$ &  $~~7.6 \pm 1.6$ &  $19.6 \pm 0.2$ &  $22.1 \pm 1.0$ &  $22.6 \pm 0.7$ &  $34.3 \pm 0.5$ &  $34.4 \pm 0.8$ &  $25.8 \pm 0.4$ &  $22.6 \pm 3.6$ &  $25.8 \pm 0.4$ \\
\midrule
\multirow{2}{*}{$0.001$} 
& MIN 
&  $~~0.0 \pm 0.0$ &  $~~0.0 \pm 0.0$ &  $~~0.0 \pm 0.0$ &   $~~8.3 \pm 0.5$ &   $~~2.0 \pm 1.9$ &   $~~8.3 \pm 0.4$ & $~~0.1 \pm 0.0$ & $~~0.1 \pm 0.0$ &  $10.5 \pm 0.4$ &  $10.7 \pm 0.6$ &  $16.7 \pm 0.6$ &  $13.8 \pm 3.5$ &  $16.5 \pm 0.6$ \\
& AVG 
&  $~~6.6 \pm 0.3$ &  $~~3.7 \pm 0.2$ &  $~~6.7 \pm 0.3$ &  $12.5 \pm 0.3$ &  $10.2 \pm 1.7$ &  $11.7 \pm 0.3$ &  $~~9.1 \pm 0.2$ &  $~~9.2 \pm 0.2$ &  $16.7 \pm 0.8$ &  $16.9 \pm 0.7$ &  $20.6 \pm 0.4$ &  $17.2 \pm 3.9$ &  $20.3 \pm 0.5$ \\
\bottomrule
\end{tabular}
}
\\
    \vspace{2em}
    \textbf{Online}\\[0.5em]
    \resizebox{\textwidth}{!}{
\begin{tabular}{p{2em}p{2em}p{4em} |
p{4em}p{4em}p{5em} |
p{4em}p{4em}p{5em} |
p{4em}p{5em} |
p{4em}p{5em} |
p{4em}p{4em}p{4em}p{0em}}
\toprule
% & Metric & Finetune & GEM & AGEM & Replay & Replay GEM & Replay AGEM \\
\vspace{-0.25em} \centering BS & \vspace{-0.25em} \centering LR & Final ACC  %
& \vspace{-0.25em} \centering Finetune & \vspace{-0.25em} \centering GEM & \vspace{-0.25em} \centering AGEM%
& \vspace{-0.25em} \centering ER & \centering ER \mbox{+ GEM} & \centering ER \mbox{+ AGEM}% 
& \vspace{-0.25em} \centering DER$^\ast$ & \centering DER$^\ast$ \mbox{+ AGEM}%
& \vspace{-0.25em} \centering BiC$^\ast$ & \centering BiC$^\ast$ \mbox{+ AGEM}%
& \centering \vspace{-0.25em} Joint & \centering Joint \mbox{+ GEM} & \centering Joint \mbox{+ AGEM} & %
\\
\midrule

%%%%%%%%%%%%%%%%%%%%%%%%%% 
% BS 10
%%%%%%%%%%%%%%%%%%%%%%%%%% 
% LR 0.1
\multirow{6}{*}{$10$} & \multirow{2}{*}{$0.1$} 
& \text{MIN} 
&  $~~0.0 \pm 0.0$ &  $~~0.0 \pm 0.0$ &  $~~0.0 \pm 0.0$ &   $~~9.0 \pm 0.4$ &  $~~0.0 \pm 0.0$ &   $~~9.1 \pm 0.3$ &  $~~0.0 \pm 0.0$ &  $~~0.0 \pm 0.0$ &   $~~7.5 \pm 0.4$ &   $~~8.6 \pm 0.2$ &  $22.1 \pm 0.5$ &  $22.9 \pm 0.6$ &   $22.8 \pm 0.4$ \\
& & \text{AVG} 
&  $~~5.0 \pm 0.4$ &  $~~1.3 \pm 0.2$ &  $~~5.0 \pm 0.4$ &  $19.3 \pm 0.4$ &  $~~6.9 \pm 0.9$ &  $19.2 \pm 0.3$ &  $~~5.5 \pm 0.3$ &  $~~5.4 \pm 0.3$ &  $15.9 \pm 0.5$ &  $17.0 \pm 0.6$ &  $38.4 \pm 1.2$ &  $38.6 \pm 0.6$ &   $38.7 \pm 0.4$ \\
\cmidrule{2-16}
% LR 0.01
& %\multirow{2}{*}{$0.01$} 
& \text{MIN} 
& $~~0.0 \pm 0.0$ & $~~0.0 \pm 0.0$ & $~~0.0 \pm 0.0$ &\cc$10.6 \pm 0.3$ &\cc$~~0.0 \pm 0.0$ &\cc$10.7 \pm 0.4$ & $~~0.0 \pm 0.0$ & $~~0.0 \pm 0.0$ & $11.1 \pm 0.4$ & $11.1 \pm 0.4$ &\cc$27.0 \pm 0.5$ &\cc$27.2 \pm 0.5$ &\cc $27.2 \pm 0.6$ \\
& \multirow{-2}{*}{$0.01$} 
& \text{AVG} 
& $~~5.1 \pm 0.3$ & $~~2.1 \pm 0.1$ & $~~5.0 \pm 0.3$ &\cc$20.7 \pm 0.3$ &\cc$~~8.5 \pm 0.9$ &\cc$20.5 \pm 0.4$ & $~~5.8 \pm 0.3$ & $~~5.9 \pm 0.2$ & $18.4 \pm 0.4$ & $18.4 \pm 0.4$ &\cc$41.2 \pm 0.6$ &\cc$41.1 \pm 0.4$ &\cc $41.4 \pm 0.2$ \\
\cmidrule{2-16}
% LR 0.001
& \multirow{2}{*}{$0.001$} 
& MIN 
&  $~~0.0 \pm 0.0$ &  $~~0.1 \pm 0.1$ &  $~~0.0 \pm 0.0$ &  $10.0 \pm 0.2$ &   $~~0.0 \pm 0.0$ &  $10.0 \pm 0.2$ &  $~~0.0 \pm 0.0$ &  $~~0.0 \pm 0.0$ &   $~~7.9 \pm 0.3$ &   $~~7.9 \pm 0.3$ &  $21.2 \pm 0.2$ &  $10.5 \pm 4.6$ &   $21.4 \pm 0.3$ \\
& & AVG 
&  $~~6.0 \pm 0.3$ &  $~~3.0 \pm 0.3$ &  $~~5.8 \pm 0.3$ &  $18.8 \pm 0.2$ &  $12.2 \pm 1.3$ &  $18.8 \pm 0.2$ &  $~~6.4 \pm 0.4$ &  $~~6.4 \pm 0.4$ &  $12.6 \pm 0.4$ &  $12.6 \pm 0.3$ &  $32.6 \pm 0.3$ &  $26.5 \pm 6.2$ &   $32.8 \pm 0.3$ \\

\midrule
%%%%%%%%%%%%%%%%%%%%%%%%%% 
% BS 64
%%%%%%%%%%%%%%%%%%%%%%%%%% 
% LR 0.1
\multirow{6}{*}{$64$} & \multirow{2}{*}{$0.1$} 
& \text{MIN  } 
&  $~~0.0 \pm 0.0$ &  $~~0.0 \pm 0.0$ &  $~~0.0 \pm 0.0$ &   $~~5.7 \pm 0.5$ &  $~~0.0 \pm 0.0$ &   $~~5.6 \pm 0.4$ &  $~~0.0 \pm 0.0$ &  $~~0.0 \pm 0.0$ &   $~~4.5 \pm 0.6$ &   $~~4.5 \pm 0.6$ &  $12.2 \pm 0.7$ &  $11.1 \pm 0.9$ &   $12.1 \pm 0.4$ \\
& & \text{AVG  } 
&  $~~2.2 \pm 0.3$ &  $~~2.4 \pm 0.8$ &  $~~2.4 \pm 0.5$ &  $15.2 \pm 0.5$ &  $~~5.5 \pm 0.3$ &  $15.4 \pm 0.4$ &  $~~3.8 \pm 0.3$ &  $~~4.0 \pm 0.3$ &  $15.2 \pm 0.4$ &  $15.2 \pm 0.4$ &  $29.1 \pm 1.0$ &  $29.2 \pm 0.9$ &   $28.8 \pm 1.0$ \\
\cmidrule{2-16}
% LR 0.01
& \multirow{2}{*}{$0.01$} 
& \text{MIN  } 
&  $~~0.0 \pm 0.0$ &  $~~0.0 \pm 0.0$ &  $~~0.0 \pm 0.0$ &   $~~8.0 \pm 0.4$ &  $~~3.3 \pm 2.1$ &   $~~8.3 \pm 0.3$ &  $~~0.0 \pm 0.0$ &  $~~0.0 \pm 0.0$ &   $~~7.9 \pm 0.3$ &   $~~7.9 \pm 0.3$ &  $17.6 \pm 0.4$ &  $18.5 \pm 0.2$ &   $18.3 \pm 0.4$ \\
& & \text{AVG  } 
&  $~~1.9 \pm 0.2$ &  $~~5.3 \pm 1.4$ &  $~~1.9 \pm 0.2$ &  $17.3 \pm 0.7$ &  $~~8.7 \pm 4.0$ &  $17.0 \pm 0.8$ &  $~~4.0 \pm 0.3$ &  $~~3.6 \pm 0.3$ &  $15.1 \pm 0.4$ &  $15.1 \pm 0.4$ &  $33.8 \pm 0.2$ &  $33.7 \pm 0.4$ &   $34.3 \pm 0.3$ \\
\cmidrule{2-16}
% LR 0.001
& \multirow{2}{*}{$0.001$} 
& MIN   
&  $~~0.0 \pm 0.0$ &  $~~0.0 \pm 0.0$ &  $~~0.0 \pm 0.0$ &   $~~6.3 \pm 0.2$ &  $~~0.0 \pm 0.0$ &   $~~6.3 \pm 0.3$ &  $~~0.0 \pm 0.0$ &  $~~0.0 \pm 0.0$ &  $~~1.8 \pm 0.3$ &  $~~1.8 \pm 0.3$ &   $~~8.2 \pm 0.3$ &   $~~1.3 \pm 0.4$ &    $~~8.3 \pm 0.3$ \\
& & AVG   
&  $~~4.1 \pm 0.2$ &  $~~6.9 \pm 1.5$ &  $~~4.2 \pm 0.3$ &  $13.0 \pm 0.5$ &  $~~5.9 \pm 1.4$ &  $13.0 \pm 0.5$ &  $~~4.1 \pm 0.3$ &  $~~4.1 \pm 0.4$ &  $~~7.0 \pm 0.4$ &  $~~7.0 \pm 0.4$ &  $16.0 \pm 0.4$ &   $~~8.4 \pm 2.4$ &   $15.7 \pm 0.4$ \\

\midrule
%%%%%%%%%%%%%%%%%%%%%%%%%% 
% BS 128
%%%%%%%%%%%%%%%%%%%%%%%%%% 
% LR 0.1
\multirow{6}{*}{$128$} & \multirow{2}{*}{$0.1$} 
& \text{MIN  } 
&  $~~0.0 \pm 0.0$ &  $~~0.0 \pm 0.0$ &  $~~0.0 \pm 0.0$ &  $~~1.5 \pm 0.2$ &  $~~0.0 \pm 0.0$ &  $~~1.6 \pm 0.3$ &  $~~0.3 \pm 0.2$ &  $~~0.0 \pm 0.0$ &   $~~2.8 \pm 0.2$ &   $~~2.8 \pm 0.2$ &   $~~3.6 \pm 0.6$ &   $~~0.9 \pm 0.8$ &    $~~4.5 \pm 0.4$ \\
& & \text{AVG  } 
&  $~~1.1 \pm 0.1$ &  $~~1.5 \pm 0.2$ &  $~~1.7 \pm 0.3$ &  $~~9.1 \pm 0.6$ &  $~~4.1 \pm 0.7$ &  $~~9.1 \pm 0.2$ &  $~~1.4 \pm 0.2$ &  $~~1.6 \pm 0.4$ &  $12.0 \pm 0.4$ &  $12.0 \pm 0.4$ &  $14.4 \pm 0.9$ &   $~~7.4 \pm 2.1$ &   $15.0 \pm 0.8$ \\
\cmidrule{2-16}
% LR 0.01
& \multirow{2}{*}{$0.01$} 
& \text{MIN  } 
&  $~~0.3 \pm 0.2$ &  $~~0.0 \pm 0.0$ &  $~~0.4 \pm 0.2$ &   $~~6.7 \pm 0.5$ &  $~~1.4 \pm 1.4$ &   $~~6.6 \pm 0.4$ &  $~~0.0 \pm 0.0$ &  $~~0.3 \pm 0.2$ &   $~~5.0 \pm 0.3$ &   $~~5.0 \pm 0.3$ &  $13.2 \pm 0.5$ &   $~~5.3 \pm 3.3$ &   $13.4 \pm 0.6$ \\
& & \text{AVG  } 
&  $~~1.3 \pm 0.1$ &  $~~5.3 \pm 0.5$ &  $~~1.6 \pm 0.2$ &  $15.2 \pm 0.2$ &  $~~6.6 \pm 2.8$ &  $15.3 \pm 0.4$ &  $~~2.1 \pm 0.1$ &  $~~2.6 \pm 0.2$ &  $12.1 \pm 0.4$ &  $12.1 \pm 0.4$ &  $25.7 \pm 1.0$ &  $11.8 \pm 5.7$ &   $25.4 \pm 0.8$ \\
\cmidrule{2-16}
% LR 0.001
& \multirow{2}{*}{$0.001$} 
& MIN   
&  $~~1.0 \pm 0.3$ &  $~~0.0 \pm 0.0$ &  $~~1.0 \pm 0.2$ &   $~~3.0 \pm 0.2$ &  $~~0.0 \pm 0.0$ &   $~~3.0 \pm 0.2$ &  $~~1.1 \pm 0.3$ &  $~~1.0 \pm 0.2$ &  $~~0.9 \pm 0.1$ &  $~~0.9 \pm 0.1$ &   $~~3.2 \pm 0.4$ &   $~~1.1 \pm 0.2$ &    $~~3.3 \pm 0.3$ \\
& & AVG   
&  $~~3.4 \pm 0.2$ &  $~~4.8 \pm 1.4$ &  $~~3.4 \pm 0.3$ &  $10.2 \pm 0.3$ &  $~~3.6 \pm 1.3$ &  $10.2 \pm 0.3$ &  $~~3.2 \pm 0.3$ &  $~~3.3 \pm 0.3$ &  $~~4.9 \pm 0.3$ &  $~~4.9 \pm 0.3$ &  $11.6 \pm 0.4$ &   $~~4.5 \pm 0.3$ &   $11.6 \pm 0.4$ \\
\bottomrule

\end{tabular}
}

    \label{apx_table:cif100}
\end{sidewaystable*}

\begin{sidewaystable*}
    %\label{apx_table:miniimg}
    \centering
    \caption{\textbf{Final average accuracy (AVG) and minimum average accuracy (MIN) for all hyperparameter settings of online and offline Split Mini-ImageNet.} Runs marked with gray background are used for comparisons in the main body. Results reported as mean $\pm$ standard error over 5 runs with different random seeds.}
    \vspace{1em}
    \textbf{Offline}\\[0.5em]
    \resizebox{\textwidth}{!}{
\begin{tabular}{p{2em}p{2em} | 
p{4em}p{4em}p{4em} |
p{4em}p{4em}p{4em} |
p{4em}p{4em} |
p{4em}p{4em} |
p{4em}p{4em}p{4em}p{0em}}
\toprule
% & Metric & Finetune & GEM & AGEM & Replay & Replay GEM & Replay AGEM \\
\vspace{-0.25em} \centering LR & Final ACC  %
& \vspace{-0.25em} \centering Finetune & \vspace{-0.25em} \centering GEM & \vspace{-0.25em} \centering AGEM%
& \vspace{-0.25em} \centering ER & \centering ER \mbox{+ GEM} & \centering ER \mbox{+ AGEM}% 
& \vspace{-0.25em} \centering DER$^\ast$ & \centering DER$^\ast$ \mbox{+ AGEM}%
& \vspace{-0.25em} \centering BiC$^\ast$ & \centering BiC$^\ast$ \mbox{+ AGEM}%
& \centering \vspace{-0.25em} Joint & \centering Joint \mbox{+ GEM} & \centering Joint \mbox{+ AGEM} & %
\\
\midrule
% LR 0.1
\multirow{2}{*}{$0.1$} 
& MIN 
&  $~~0.0 \pm 0.0$ &  $~~0.0 \pm 0.0$ &  $~~0.0 \pm 0.0$ & \cc $~~8.6 \pm 0.5$ & \cc $~~0.0 \pm 0.0$ & \cc $~~8.3 \pm 0.4$ &  $~~0.0 \pm 0.0$ &  $~~0.0 \pm 0.0$ & $~~7.8 \pm 0.9$ & $~~8.1 \pm 0.8$ & \cc $16.5 \pm 0.5$ & \cc  $~~0.0 \pm 0.0$ & \cc $16.4 \pm 0.4$ \\
& AVG 
&  $~~3.2 \pm 0.1$ &  $~~1.0 \pm 0.0$ &  $~~3.2 \pm 0.1$ & \cc $16.9 \pm 0.9$ & \cc $~~4.7 \pm 1.6$ & \cc $16.9 \pm 0.5$ &  $~~4.3 \pm 0.2$ &  $~~4.2 \pm 0.8$ &  $20.3 \pm 0.6$ &  $20.4 \pm 1.2$ & \cc $28.1 \pm 0.6$ & \cc $~~3.3 \pm 0.4$ & \cc $28.5 \pm 0.7$ \\
\midrule
\multirow{2}{*}{$0.01$} 
& MIN 
&  $~~0.0 \pm 0.0$ &  $~~0.0 \pm 0.0$ &  $~~0.0 \pm 0.0$ &   $~~4.1 \pm 0.5$ &  $~~0.0 \pm 0.0$ &   $~~4.0 \pm 0.7$ &  \cc $~~0.1 \pm 0.0$ & \cc $~~0.2 \pm 0.1$ & \cc $13.3 \pm 0.3$ & \cc $13.5 \pm 0.2$ &   $~~8.9 \pm 1.1$ &  $~~0.0 \pm 0.0$ &   $~~8.5 \pm 0.7$ \\
& AVG 
&  $~~3.1 \pm 0.1$ &  $~~1.9 \pm 0.1$ &  $~~3.1 \pm 0.1$ &  $13.5 \pm 0.6$ &  
$~~6.4 \pm 1.9$ &  $14.1 \pm 0.5$ & \cc $~~7.4 \pm 0.5$ & \cc $~~7.6 \pm 0.7$ & \cc $25.9 \pm 0.7$ & \cc $26.2 \pm 0.5$ &  $19.1 \pm 1.2$ &  $~~4.6 \pm 0.5$ &  $18.5 \pm 0.9$ \\
\midrule
\multirow{2}{*}{$0.001$} 
& MIN 
&  $~~0.0 \pm 0.0$ &  $~~0.0 \pm 0.0$ &  $~~0.0 \pm 0.0$ &   $~~4.0 \pm 0.3$ &  $~~0.0 \pm 0.0$ &  $~~3.8 \pm 0.3$ &  $~~0.0 \pm 0.0$ &  $~~0.0 \pm 0.0$ &   $~~6.8 \pm 0.6$ &   $~~6.7 \pm 0.4$ &   $~~9.6 \pm 0.8$ &  $~~0.0 \pm 0.0$ &   $~~9.6 \pm 0.7$ \\
& AVG 
&  $~~3.2 \pm 0.1$ &  $~~3.6 \pm 0.6$ &  $~~3.4 \pm 0.1$ &  $10.1 \pm 0.4$ &  $~~5.2 \pm 0.8$ &  $~~9.9 \pm 0.2$ &  $~~4.2 \pm 0.2$ &  $~~4.5 \pm 0.2$ &  $14.4 \pm 0.4$ &  $14.2 \pm 0.3$ &  $16.7 \pm 0.7$ &  $~~6.4 \pm 1.1$ &  $16.9 \pm 0.7$ \\
\bottomrule
\end{tabular}
}
\\
    \vspace{2em}
    \textbf{Online}\\[0.5em]
    \resizebox{\textwidth}{!}{
\begin{tabular}{p{2em}p{2em}p{2em} | 
p{4em}p{4em}p{4em} |
p{4em}p{4em}p{4em} | 
p{4em}p{4em} | 
p{4em}p{4em} | 
p{4em}p{4em}p{4em}p{0em}}
\toprule
% & Metric & Finetune & GEM & AGEM & Replay & Replay GEM & Replay AGEM \\
\vspace{-0.25em} \centering BS & \vspace{-0.25em} \centering LR & Final ACC  %
& \vspace{-0.25em} \centering Finetune & \vspace{-0.25em} \centering GEM & \vspace{-0.25em} \centering AGEM%
& \vspace{-0.25em} \centering ER & \centering ER \mbox{+ GEM} & \centering ER \mbox{+ AGEM}% 
& \vspace{-0.25em} \centering DER$^\ast$ & \centering DER$^\ast$ \mbox{+ AGEM}%
& \vspace{-0.25em} \centering BiC$^\ast$ & \centering BiC$^\ast$ \mbox{+ AGEM}%
& \centering \vspace{-0.25em} Joint & \centering Joint \mbox{+ GEM} & \centering Joint \mbox{+ AGEM} & %
\\
\midrule

%%%%%%%%%%%%%%%%%%%%%%%%%% 
% BS 10
%%%%%%%%%%%%%%%%%%%%%%%%%% 
% LR 0.1
\multirow{6}{*}{$10$} & \multirow{2}{*}{$0.1$} 
& \text{MIN} 
&  $~~0.0 \pm 0.0$ &  $~~0.0 \pm 0.0$ &  $~~0.0 \pm 0.0$ &  $~~0.3 \pm 0.1$ &  $~~0.0 \pm 0.0$ &  $~~0.3 \pm 0.2$ &  $~~0.8 \pm 0.2$ &  $~~1.1 \pm 0.0$ &  $~~0.6 \pm 0.2$ &  $~~0.1 \pm 0.1$ &  $~~0.2 \pm 0.1$ &   $0.1 \pm 0.0$ & $~~1.0 \pm 0.3$ \\
& & \text{AVG} 
 &  $~~1.0 \pm 0.0$ &  $~~1.6 \pm 0.5$ &  $~~1.1 \pm 0.1$ &  $~~6.6 \pm 0.5$ &  $~~1.8 \pm 0.5$ &  $~~6.4 \pm 0.6$ &  $~~1.0 \pm 0.0$ &  $~~1.0 \pm 0.0$ &  $~~8.3 \pm 0.8$ &  $~~3.7 \pm 0.3$ &  $~~3.9 \pm 0.4$ & $~~3.4 \pm 0.4$ & $~~8.0 \pm 1.1$ \\
\cmidrule{2-16}
% LR 0.01
& \multirow{2}{*}{$0.01$} 
& \text{MIN} 
&  $~~0.1 \pm 0.1$ &  $~~0.0 \pm 0.0$ &  $~~0.3 \pm 0.2$ &  $~~1.0 \pm 0.2$ &  $~~0.0 \pm 0.0$ &   $~~0.9 \pm 0.3$ &  $~~0.4 \pm 0.2$ &  $~~0.2 \pm 0.2$ &   $~~1.6 \pm 0.2$ &  $~~0.3 \pm 0.1$ &  $~~0.5 \pm 0.2$ &   $~~0.0 \pm 0.0$ &    $~~2.0 \pm 0.2$ \\
& & \text{AVG} 
&  $~~1.1 \pm 0.1$ &  $~~3.0 \pm 0.5$ &  $~~1.3 \pm 0.1$ &  $10.8 \pm 0.4$ &  $~~5.5 \pm 0.6$ &  $10.9 \pm 0.7$ &  $~~1.5 \pm 0.2$ &  $~~1.7 \pm 0.3$ &  $16.2 \pm 0.6$ &  $~~7.8 \pm 0.5$ &  $~~8.0 \pm 0.6$ &   $~~4.8 \pm 0.6$ &   $16.7 \pm 0.9$ \\
\cmidrule{2-16}
% LR 0.001
& \multirow{2}{*}{$0.001$} 
& MIN 
&  $~~0.2 \pm 0.1$ &  $~~0.0 \pm 0.0$ &  $~~0.2 \pm 0.1$ &   $~~1.4 \pm 0.3$ &  $~~0.0 \pm 0.0$ &   $~~1.5 \pm 0.3$ &  $~~0.4 \pm 0.1$ &  $~~0.6 \pm 0.2$ &   $~~1.3 \pm 0.2$ &  $~~0.0 \pm 0.0$ &  $~~0.0 \pm 0.0$ &   $~~0.0 \pm 0.0$ &    $~~1.5 \pm 0.3$ \\
& & AVG 
&  $~~1.2 \pm 0.2$ &  $~~4.0 \pm 0.8$ &  $~~1.2 \pm 0.1$ &  $11.0 \pm 0.3$ &  $~~5.2 \pm 0.9$ &  $11.1 \pm 0.4$ &  $~~1.8 \pm 0.2$ &  $~~2.5 \pm 0.2$ &  $12.3 \pm 0.3$ &  $~~4.5 \pm 0.3$ &  $~~4.5 \pm 0.3$ &   $~~5.4 \pm 0.9$ &   $12.3 \pm 0.3$ \\

\midrule
%%%%%%%%%%%%%%%%%%%%%%%%%% 
% BS 64
%%%%%%%%%%%%%%%%%%%%%%%%%% 
% LR 0.1
\multirow{6}{*}{$64$} & \multirow{2}{*}{$0.1$} 
& \text{MIN  } 
&  $~~0.0 \pm 0.0$ &  $~~0.0 \pm 0.0$ &  $~~0.0 \pm 0.0$ &  $~~0.7 \pm 0.2$ &  $~~0.0 \pm 0.0$ &  $~~1.0 \pm 0.2$ &  $~~0.4 \pm 0.3$ &  $~~0.4 \pm 0.3$ &  $~~0.2 \pm 0.1$ &  $~~0.2 \pm 0.1$ &   $~~3.1 \pm 0.4$ &   $~~0.0 \pm 0.0$ &    $~~3.1 \pm 0.6$ \\
& & \text{AVG  } 
&  $~~1.0 \pm 0.0$ &  $~~1.0 \pm 0.1$ &  $~~1.3 \pm 0.1$ &  $~~8.3 \pm 0.3$ &  $~~1.7 \pm 0.2$ &  $~~8.6 \pm 0.6$ &  $~~1.1 \pm 0.1$ &  $~~1.1 \pm 0.1$ &  $~~5.0 \pm 0.5$ &  $~~5.3 \pm 0.4$ &  $16.9 \pm 0.8$ &   $~~3.5 \pm 0.5$ &   $16.8 \pm 1.4$ \\
\cmidrule{2-16}
% LR 0.01
& \multirow{2}{*}{$0.01$} 
& \text{MIN  } 
& $~~0.0 \pm 0.0$ &  $~~0.0 \pm 0.0$ &  $~~0.0 \pm 0.0$ & \cc $~~3.3 \pm 0.2$ & \cc $~~0.0 \pm 0.0$ & \cc $~~3.2 \pm 0.2$ &  $~~0.1 \pm 0.1$ &  $~~0.1 \pm 0.0$ &   $~~2.0 \pm 0.3$ &   $~~1.9 \pm 0.4$ & \cc $~~9.1 \pm 0.2$ & \cc $~~0.0 \pm 0.0$ & \cc $~~8.9 \pm 0.3$ \\
& & \text{AVG  } 
&  $~~1.2 \pm 0.1$ &  $~~2.7 \pm 0.9$ &  $~~1.4 \pm 0.2$ & \cc $13.9 \pm 0.5$ & \cc $~~5.8 \pm 0.3$ & \cc $14.6 \pm 0.3$ &  $~~1.6 \pm 0.3$ &  $~~1.8 \pm 0.2$ &  $11.1 \pm 0.5$ &  $11.0 \pm 0.7$ &\cc $27.9 \pm 0.8$ & \cc $~~5.5 \pm 0.5$ & \cc $28.3 \pm 1.3$ \\
\cmidrule{2-16}
% LR 0.001
& \multirow{2}{*}{$0.001$} 
& MIN   
&  $~~0.2 \pm 0.1$ &  $~~0.0 \pm 0.0$ &  $~~0.3 \pm 0.3$ &   $~~3.7 \pm 0.5$ &  $~~0.0 \pm 0.0$ &   $~~3.8 \pm 0.5$ &  $~~0.9 \pm 0.1$ &  $~~0.9 \pm 0.3$ &  $~~0.8 \pm 0.1$ &  $~~0.8 \pm 0.2$ &   $~~4.1 \pm 1.5$ &   $~~0.0 \pm 0.0$ &    $~~5.9 \pm 0.5$ \\
& & AVG   
&  $~~1.4 \pm 0.1$ &  $~~8.3 \pm 0.8$ &  $~~1.7 \pm 0.2$ &  $13.6 \pm 0.2$ &  $~~6.6 \pm 0.9$ &  $13.7 \pm 0.3$ &  $~~2.3 \pm 0.2$ &  $~~1.8 \pm 0.1$ &  $~~6.3 \pm 0.3$ &  $~~6.4 \pm 0.3$ &  $12.4 \pm 3.6$ &   $~~3.6 \pm 1.1$ &   $17.9 \pm 0.5$ \\

\midrule
%%%%%%%%%%%%%%%%%%%%%%%%%% 
% BS 128
%%%%%%%%%%%%%%%%%%%%%%%%%% 
% LR 0.1
\multirow{6}{*}{$128$} & \multirow{2}{*}{$0.1$} 
& \text{MIN  } 
&  $~~0.0 \pm 0.0$ &  $~~0.0 \pm 0.0$ &  $~~0.0 \pm 0.0$ &  $~~0.3 \pm 0.1$ &  $~~0.0 \pm 0.0$ &  $~~0.3 \pm 0.2$ &  $~~0.8 \pm 0.2$ &  $~~1.1 \pm 0.0$ &  $~~0.1 \pm 0.1$ &  $~~0.2 \pm 0.1$ &  $~~0.6 \pm 0.2$ &   $~~0.1 \pm 0.0$ &    $~~1.0 \pm 0.3$ \\
& & \text{AVG  } 
&  $~~1.0 \pm 0.0$ &  $~~1.6 \pm 0.5$ &  $~~1.1 \pm 0.1$ &  $~~6.6 \pm 0.5$ &  $~~1.8 \pm 0.5$ &  $~~6.4 \pm 0.6$ &  $~~1.0 \pm 0.0$ &  $~~1.0 \pm 0.0$ &  $~~3.7 \pm 0.3$ &  $~~3.9 \pm 0.4$ &  $~~8.3 \pm 0.8$ &   $~~3.4 \pm 0.4$ &    $~~8.0 \pm 1.1$ \\
\cmidrule{2-16}
% LR 0.01
& \multirow{2}{*}{$0.01$} 
& \text{MIN  } 
&  $~~0.1 \pm 0.1$ &  $~~0.0 \pm 0.0$ &  $~~0.3 \pm 0.2$ &   $~~1.0 \pm 0.2$ &  $~~0.0 \pm 0.0$ &   $~~0.9 \pm 0.3$ &  $~~0.4 \pm 0.2$ &  $~~0.2 \pm 0.2$ &  $~~0.3 \pm 0.1$ &  $~~0.5 \pm 0.2$ &   $~~1.6 \pm 0.2$ &   $~~0.0 \pm 0.0$ &    $~~2.0 \pm 0.2$ \\
& & \text{AVG  } 
&  $~~1.1 \pm 0.1$ &  $~~3.0 \pm 0.5$ &  $~~1.3 \pm 0.1$ &  $10.8 \pm 0.4$ &  $~~5.5 \pm 0.6$ &  $10.9 \pm 0.7$ &  $~~1.5 \pm 0.2$ &  $~~1.7 \pm 0.3$ &  $~~7.8 \pm 0.5$ &  $~~8.0 \pm 0.6$ &  $16.2 \pm 0.6$ &   $~~4.8 \pm 0.6$ &   $16.7 \pm 0.9$ \\
\cmidrule{2-16}
% LR 0.001
& \multirow{2}{*}{$0.001$} 
& MIN   
&  $~~0.2 \pm 0.1$ &  $~~0.0 \pm 0.0$ &  $~~0.2 \pm 0.1$ &   $~~1.4 \pm 0.3$ &  $~~0.0 \pm 0.0$ &   $~~1.5 \pm 0.3$ &  $~~0.4 \pm 0.1$ &  $~~0.6 \pm 0.2$ &  $~~0.0 \pm 0.0$ &  $~~0.0 \pm 0.0$ &   $~~1.3 \pm 0.2$ &   $~~0.0 \pm 0.0$ &    $~~1.5 \pm 0.3$ \\
& & AVG   
&  $~~1.2 \pm 0.2$ &  $~~4.0 \pm 0.8$ &  $~~1.2 \pm 0.1$ &  $11.0 \pm 0.3$ &  $~~5.2 \pm 0.9$ &  $11.1 \pm 0.4$ &  $~~1.8 \pm 0.2$ &  $~~2.5 \pm 0.2$ &  $~~4.5 \pm 0.3$ &  $~~4.5 \pm 0.3$ &  $12.3 \pm 0.3$ &   $~~5.4 \pm 0.9$ &   $12.3 \pm 0.3$ \\
\bottomrule

\end{tabular}
}

    \label{apx_table:miniimg}
\end{sidewaystable*}

\end{document}